\documentclass[10pt,twocolumn,letterpaper]{article}

\usepackage[pagenumbers]{iccv} %

\usepackage{footnote}
\usepackage{multirow}
\usepackage{makecell} %
\usepackage{threeparttable}

\usepackage{caption}
\usepackage{graphicx} 
\usepackage{colortbl}
\usepackage{graphicx}
\usepackage{multicol}
\usepackage{multirow}
\usepackage{makecell}
\usepackage{tablefootnote}
\usepackage{amsmath}
\usepackage{amssymb}
\usepackage{tabularx}
\usepackage[dvipsnames]{xcolor}

\usepackage{siunitx} %
\usepackage{colortbl}
\usepackage{graphicx}
\usepackage{lipsum}
\usepackage{nicefrac}
\usepackage{tabularray}

\usepackage{algorithm}%
\usepackage{algorithmicx}%
\usepackage{algpseudocode}

\def\eg{\emph{e.g.}\xspace} 
\def\ie{\emph{i.e.}\xspace} 

\def\wrt{\emph{w.r.t.}\xspace}

\def\sota{state-of-the-art\xspace}

\def\scannet{ScanNet\xspace}
\def\scannetpp{ScanNet++\xspace}
\def\matterport{Matterport3D\xspace}

\def\arkit{ARKitScenes\xspace}
\def\replica{Replica\xspace}

\newcolumntype{R}{>{\raggedleft\arraybackslash}X} %
\newcolumntype{C}{>{\centering\arraybackslash}X} %

\newcommand{\ours}{SceneSplat\xspace}
\newcommand{\ourdata}{SceneSplat-7K\xspace}
\newcommand{\boldparagraph}[1]{\vspace{0.1em}\noindent{\bf #1}}

\newcommand{\notoc}[1]{%
    \addtocontents{toc}{\protect\setcounter{tocdepth}{-1}}%
    #1%
    \addtocontents{toc}{\protect\setcounter{tocdepth}{2}}%
}

\newcommand\blfootnote[1]{%
  \begingroup
  \renewcommand\thefootnote{}\footnote{#1}%
  \addtocounter{footnote}{-1}%
  \endgroup
}

\definecolor{iccvblue}{rgb}{0.21,0.49,0.74}
\usepackage[pagebackref,breaklinks,colorlinks,allcolors=iccvblue]{hyperref}

\title{SceneSplat: Gaussian Splatting-based Scene Understanding\\ with Vision-Language Pretraining}

\author{  
\thanks{Indicates equal contribution.}~Yue Li\textsuperscript{1}, 
\footnotemark[1]~Qi Ma\textsuperscript{2,3}, 
Runyi Yang\textsuperscript{3},
Huapeng Li\textsuperscript{2},
Mengjiao Ma\textsuperscript{3,4},
\footnotemark[2]~~Bin Ren\textsuperscript{3,5,6},
Nikola Popovic\textsuperscript{3} \\
Nicu Sebe\textsuperscript{6}, 
Ender Konukoglu\textsuperscript{2}, 
Theo Gevers\textsuperscript{1}, 
Luc Van Gool\textsuperscript{2,3},
Martin R. Oswald\textsuperscript{1},
Danda Pani Paudel\textsuperscript{3}\\
\normalsize{
\textsuperscript{1}University of Amsterdam\quad 
\textsuperscript{2}Computer Vision Lab, ETH Zurich\quad 
\textsuperscript{3}INSAIT, Sofia University "St. Kliment Ohridski"} \\
\normalsize{
\textsuperscript{4}Nanjing University of Aeronautics and Astronautics\quad 
\textsuperscript{5}University of Pisa\quad 
\textsuperscript{6}University of Trento} \\
}

\begin{document}

\twocolumn[{%
  \renewcommand\twocolumn[1][]{#1}%
  \maketitle
  \begin{center}
  \vspace{-10mm}
  \includegraphics[width=0.85\linewidth, trim=0 0 0pt 0, clip]{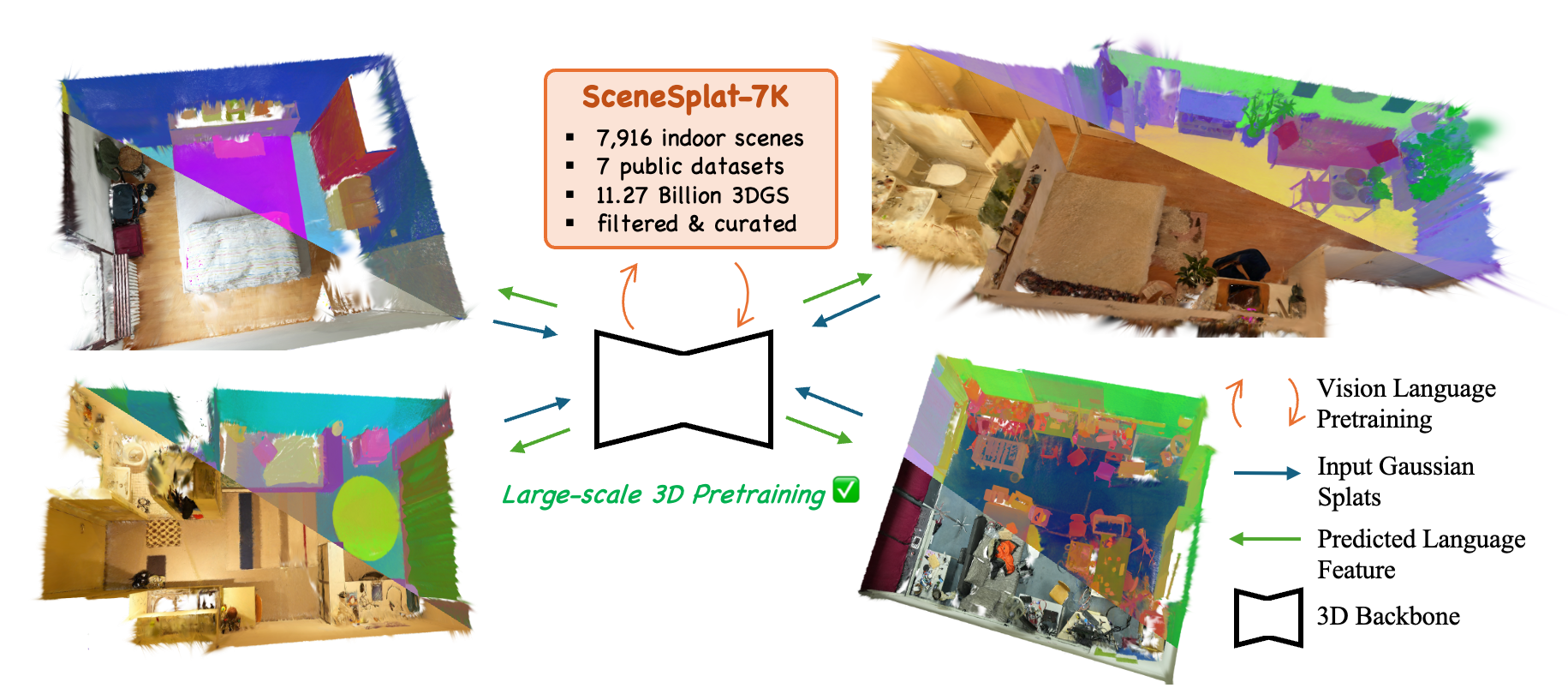}

  \end{center}
  \vspace{-15pt}
  \captionof{figure}{
        We present the 3DGS indoor dataset \textbf{\ourdata} which includes 7K scenes generated from ARKitScenes~\cite{baruch2021arkitscenes}, Replica~\cite{straub2019replica}, ScanNet~\cite{dai2017scannet}, ScanNet++~\cite{yeshwanth2023scannet++}, Hypersim\cite{roberts2021hypersim}, 3RScan~\cite{wald2019rio}, and Matterport3D~\cite{chang2017matterport3d}. Leveraging this high-quality dataset, we propose \textbf{\ours}, the first model  to predict open-vocabulary language features for millions of 3D Gaussians in a single forward pass. 
    }
    \label{fig:teaser}
    \vspace{1.9em}
}]

\blfootnote{%
$\ast$ indicates equal contribution. %
$\dagger$ indicates the corresponding author: Bin Ren %
\textless\href{mailto:bin.ren@insait.ai}{bin.ren@insait.ai}\textgreater.%
}

\notoc{\begin{abstract}
Recognizing arbitrary or previously unseen categories is essential for comprehensive real-world 3D scene understanding. Currently, all existing methods rely on 2D or textual modalities during training or together at inference. This highlights the clear absence of a model capable of processing 3D data alone for learning semantics end-to-end, along with the necessary data to train such a model. Meanwhile, 3D Gaussian Splatting (3DGS) has emerged as the de facto standard for 3D scene representation across various vision tasks. However, effectively integrating semantic reasoning into 3DGS in a generalizable manner remains an open challenge.
To address these limitations, we introduce SceneSplat in \cref{fig:teaser}, to our knowledge the first large-scale 3D indoor scene understanding approach that operates natively on 3DGS. Furthermore, we propose a self-supervised learning scheme that unlocks rich 3D feature learning from unlabeled scenes. 
To power the proposed methods, we introduce SceneSplat-7K, the first large-scale 3DGS dataset for indoor scenes, comprising 7916 scenes derived from seven established datasets, such as ScanNet and Matterport3D. Generating SceneSplat-7K required computational resources equivalent to 150 GPU days on an L4 GPU, enabling standardized benchmarking for 3DGS-based reasoning for indoor scenes.
Our exhaustive experiments on SceneSplat-7K demonstrate the significant benefit of the proposed method over the established baselines. Our code, model, and datasets will be released at \href{https://github.com/unique1i/SceneSplat}{SceneSplat}.

\vspace{-2mm}

\end{abstract}

}    
\notoc{\section{Introduction}
\label{sec:intro}

The ability to interpret arbitrary queries rather than being limited to a closed set of categories is crucial for 3D understanding models to generalize across diverse real-world scenarios.
Traditional 3D vision systems are typically trained on fixed closed-set category labels drawn from datasets like ScanNet~\cite{dai2017scannet}. Such label spaces fail to capture the diversity of concepts in real-world environments. This gap has spurred significant interest in open-vocabulary 3D scene understanding, which aims to recognize arbitrary or unseen categories beyond a predefined taxonomy~\cite{ding2022pla,yang2024regionplc,jiang2024open}. Achieving this capability would empower the model to reason about novel objects in a scene using natural language descriptors.

Achieving open-vocabulary recognition in 3D is challenging due to the scarcity of large-scale 3D-text paired data. Breakthroughs in 2D vision, driven by internet-scale image-text pre-training, cannot be directly leveraged in 3D due to the absence of analogous 3D datasets with rich textual annotations.
To address this gap, current methods resort to multi-modality fusion, distilling knowledge from 2D vision-language models into 3D data. Some approaches project 3D points into source images and use a pretrained 2D backbone (\eg, CLIP) to supervise 3D feature learning~\cite{zhang2022pointclip, peng2023openscene, zhu2023pointclip}; others generate synthetic captions for 3D scenes to explicitly associate point clouds with semantic-rich text~\cite{rozenberszki2022language, Ding_2023_CVPR}. 
However, all current methods rely on 2D or textual modalities during training, or together at inference, to compensate for limited 3D semantics. 
This highlights a key limitation: 
\textit{the absence of a robust model for processing 3D data end-to-end for semantic learning}, along with 
\textit{the lack of sufficient data for training such a model}.

The field of 3D representation is rapidly evolving. While classical 3D networks rely on point clouds or voxel grids~\cite{yang2023swin3d, wu2022point, wu2024point}, recent approaches inspired by radiance fields have greatly improved scene representation. Notably, 3D Gaussian Splatting (3DGS)~\cite{kerbl20233d} has emerged as an efficient representation, achieving state-of-the-art performance in view synthesis and geometry modeling~\cite{yu2024mip, zhang2024rade}.
Unlike discrete point clouds, 3DGS provides a compact formulation that can be optimized per scene for 2D synthesis, while the underlying Gaussians also softly encode rich 3D structures (positions, shapes, and opacities of the regions). 
This makes it a unique candidate for 3D scene cues, as it naturally fuses geometry and appearance information. 
However, integrating semantic reasoning into 3DGS is nontrivial. The naïve way of optimizing additional semantic features in 3DGS~\cite{kerr2023lerf} is inefficient and limited to a single scene. 
Consequently, generalizable open-vocabulary understanding in 3DGS remains unexplored, as no existing model directly processes 3D Gaussian parameters. Moreover, no large-scale scene-level 3DGS dataset is available to train such models, further limiting the progress in this direction.

We address these limitations with \ours, to our knowledge the first large-scale 3D indoor scene understanding approach that operates natively on 3D Gaussian splats. The proposed model is powered by \ourdata, a carefully curated 3DGS indoor scene dataset spanning around 7K scenes. \ours introduces a 3DGS encoder that takes as input the parameters of a Gaussian-splat scene (center, scale, color, opacity) and outputs semantic features in a per-primitive manner, in a single forward pass. We leverage supervision from vision-language models to train SceneSplat’s encoder to produce CLIP-aligned embeddings, effectively bridging language and 3D without explicit 2D fusion at runtime. Furthermore, we propose GaussSSL, a self-supervised learning scheme that unlocks rich 3D feature learning from unlabeled scenes. GaussSSL operates through three synergistic strategies: Masked Gaussian Modeling (MGM), self-distillation, and optional Language-Gaussian Feature Alignment, allowing the model to separate high-level semantic signals from raw parameters alone. This self-supervised pretraining is fueled by large amounts of 3DGS scenes in the \ourdata dataset. Our contributions can be summarized as follows: 

$\bullet$ We present \ourdata, a high-quality large-scale Gaussian splats dataset spanning 7K indoor scenes, which boosts 3DGS scene understanding research.

$\bullet$ We propose \ours, a model that unlocks open-vocabulary recognition for 3D Gaussian splats and achieves \sota zero-shot semantic segmentation performance on three fine-grained benchmarks.

$\bullet$ We incorporate annotation-free self-supervised training mechanisms on this large-scale indoor dataset and demonstrate their effectiveness in the downstream indoor segmentation task.
}
\notoc{\section{Related Work}
\label{sec:related-work}
\begin{table*}[!t]
    \centering
    \resizebox{\textwidth}{!}{
        \setlength{\extrarowheight}{0.2pt}
        \setlength{\tabcolsep}{2pt}
        \begin{tabular}{l|cccccccc|>{\columncolor{gray!15}}c}
            \toprule[0.95pt]
        \textbf{Metric} & \textbf{ScanNet\cite{dai2017scannet}} & \textbf{ScanNet++\cite{yeshwanth2023scannet++}} & \textbf{ScanNet++v2\cite{yeshwanth2023scannet++}} & \textbf{Replica\cite{straub2019replica}} & \textbf{Hypersim\cite{roberts2021hypersim}} & \textbf{3RScan\cite{wald2019rio}} & \textbf{ARKitScenes\cite{baruch2021arkitscenes}} & \textbf{Matterport3D\cite{chang2017matterport3d}} & \textbf{Scenesplat-7K} \\
        \midrule[0.6pt]
        \textbf{Raw Scenes} & 1613 & 380 & 1006 & 8 & 461 & 1482(scans) & 1970 & 2194(regions) & 9114 \\
        \textbf{GS Scenes} & 1613 & 330 & 956 & 8 & 448 & 632(scans) & 1947 & 1982(regions) & 7916 \\
        \textbf{RGB Frames } & 2.5M & 228K & 1.1M & 16K & 77K & 156K & 450K & 194K & 4.72M \\

        \textbf{Storage} & 600GB & 152GB & 447GB & 7GB & 251GB & 235GB & 577GB & 492GB & 2.76TB \\
        
        \midrule[0.6pt]
        \textbf{PSNR} & 29.07 & 29.49 & 29.11 & 41.25 & 25.93 & 27.46 & 29.18 & 32.34 & 29.64 \\
        \textbf{Depth Loss} & 0.031 & 0.019 & 0.015 & 0.002 & 0.228 & 0.018 & 0.0131 & 0.033 & 0.035\\
        \textbf{SSIM} & 0.869 & 0.924 & 0.933 & 0.980 & 0.894 & 0.881 & 0.885 & 0.916 & 0.897 \\
        \textbf{LPIPS} & 0.236 & 0.133 & 0.116 & 0.0396 & 0.157 & 0.335 & 0.294 & 0.145 & 0.212 \\
        \midrule[0.6pt]

         \textbf{GS per scene} & 1.50M & 1.56M & 1.89M & 1.50M & 2.84M & 1.50M & 1.19M & 1.0M & 1.42M \\
         \textbf{Total GS} & 2419.5M & 513.4M & 1810.3M & 12.0M & 1,237.5M & 948.0M & 2,316.9M & 1,982M & 11.27B \\
        \textbf{GPU Time (L4)} & 593h & 177h & 594h & 4h & 176h & 576h & 811h & 661h & \cellcolor{gray!20}3592 h \\
        \bottomrule[0.95pt]
        \end{tabular}
        } 
        \vspace{-2mm}
    \caption{
            \textbf{Dataset Statistics.} The proposed SceneSplat-7K dataset includes various 3D Gaussian Splatting datasets generated from \scannet\cite{dai2017scannet}, \scannetpp \cite{yeshwanth2023scannet++}, \scannetpp v2, Replica\cite{straub2019replica}, Hypersim\cite{roberts2021hypersim}, 3RScan\cite{wald2019rio}, ARKitScenes\cite{baruch2021arkitscenes}, and Matterport3D\cite{chang2017matterport3d}. The dataset contains \textbf{7,916 scenes} and \textbf{11.27 Billion 3DGS}. Constructing this dataset required computational resources equivalent to \textbf{150 GPU-days} on one NVIDIA L4 GPU. SceneSplat-7K achieves high-fidelity reconstruction quality with an average PSNR \textbf{29.64 dB}.}%

        \label{tab:dataset}
    \vspace{-3mm}
\end{table*}

\noindent\textbf{3D Indoor Datasets.} 
The advancement of 3D deep learning has driven the development of various indoor datasets for scene understanding, reconstruction, and representation learning~\cite{dai2017scannet, yeshwanth2023scannet++, roberts2021hypersim, baruch2021arkitscenes, straub2019replica, wald2019rio, chang2017matterport3d}. 
ScanNet~\cite{dai2017scannet} serves as a fundamental benchmark with 1,513 real-world indoor scenes and dense annotations, while ScanNet++~\cite{yeshwanth2023scannet++} extends it with additional real and synthetic scenes. 
Hypersim~\cite{roberts2021hypersim} offers photorealistic synthetic environments, and ARKitScenes~\cite{baruch2021arkitscenes} captures 1,661 real-world indoor scenes using ARKit. 
Replica~\cite{straub2019replica} provides high-fidelity reconstructions, and 3RScan~\cite{wald2019rio} supports multi-view analysis with 1,482 scans. Habitat-Matterport3D~\cite{chang2017matterport3d} integrates Matterport3D into an embodied AI simulation platform.
These datasets are essential for 3D perception but lack large-scale support for emerging 3D representations like 3DGS. 
Prior works \cite{irshad2024nerfmae, ma2024implicitzoolargescaledatasetneural} have explored NeRF-based representations, and ShapeSplat~\cite{ma2024shapesplat} introduced Gaussian-splatted objects, but no dataset currently supports indoor scene understanding with Gaussian Splatting. To address this limitation, we introduce a 3D Gaussian Splatting scene dataset derived from 7 widely used indoor datasets\cite{dai2017scannet, yeshwanth2023scannet++, roberts2021hypersim, baruch2021arkitscenes, straub2019replica, wald2019rio, chang2017matterport3d}. Our dataset offers a rich collection of indoor scenes from both real-world and synthetic sources, featuring high-quality Gaussian splats, facilitating the transition from object-level to scene-level Gaussian splatting and advancing large-scale 3D scene understanding.

\noindent\textbf{Open Vocabulary Scene Understanding.}
Recent advancements in open-vocabulary models have significantly expanded the capabilities of vision-language understanding. Foundation models such as DINO~\cite{zhang2022dino, oquab2023dinov2} have enabled self-supervised visual feature extraction at scale, effectively supporting tasks like detection and segmentation. SAM~\cite{kirillov2023segment, ravi2024sam} introduced the capability of prompt-driven segmentation, generalizing robustly across diverse datasets and tasks. CLIP~\cite{radford2021learning} pioneered aligning visual and textual embedding spaces, allowing for zero-shot transfer across numerous downstream tasks. 
Subsequently, SigLIP \cite{zhai2023sigmoid, tschannen2025siglip} improved alignment by introducing non-linear activation techniques, enhancing open-vocabulary performance marginally. Recent works~\cite{zhou2024feature, zheng2024gaussiangrasper, qin2024langsplat, cheng2024occam, peng2023openscene, guo2024semantic} have also introduced 2D open-vocabulary foundation models into 3D by utilizing NeRF~\cite{mildenhall2021nerf} or 3DGS~\cite{kerbl20233d}. LERF~\cite{kerr2023lerf, rashid2023language} integrated language queries into NeRF-based models, enabling rich semantic querying within 3D scenes. LangSplat~\cite{qin2024langsplat} combined 3DGS with open-vocabulary embeddings from SAM~\cite{ravi2024sam} and CLIP~\cite{radford2021learning}, facilitating open-vocabulary semantic scene understanding. OccamLGS \cite{cheng2024occam} further optimized this process by employing feature lifting, directly projecting all 2D features into 3DGS, and reducing computation time from hours to seconds. 
However, these methods require time-consuming preprocessing of images using 2D foundation models. 
In contrast, we propose a pipeline and dataset for training a 3D foundation model, enabling feed-forward open-vocabulary understanding of 3DGS.

\noindent\textbf{3D Representation Learning.} 
Representation learning in 3D, akin to its success in the 2D image domain, is crucial for extracting meaningful features from 3D data. Previous approaches have focused on either architectural advancements or learning paradigms. 
Many methods have sought to map 3D data into 2D patches, leveraging conventional 2D CNNs or Vision Transformers~\cite{vaswani2017attention,dosovitskiy2020image,ren2023masked}. Although promising, this approach often overlooks the inherent properties of 3D data due to the embedding strategy. 
To address this, research has shifted toward specialized architectures (\ie, PointCNN~\cite{li2018pointcnn}, PointNet~\cite{qi2017pointnet}, Point Transformer~\cite{zhao2021point, wu2022point, wu2024point}, and Pointmamba~\cite{liang2025pointmamba}).
However, labeling 3D data requires extensive effort and domain expertise, making it quite challenging. This has highlighted the importance of self-supervised learning (SSL) for 3D data, with methods typically falling into contrastive or generative categories. Contrastive methods aim to differentiate similar and dissimilar examples, but often suffer from issues such as overfitting and mode collapse, exacerbated by the sparse nature of 3D data~\cite{qi2023contrast,ren2024bringing}. 
On the other hand, generative methods, inspired by BERT's \textit{mask-and-reconstruct} strategy~\cite{devlin2018bert}, have proven more effective. Masked Autoencoders~\cite{he2022masked}, originally designed for 2D images, have been adapted for 3D data types, including point clouds~\cite{PointMAE,zhang2022point}, meshes~\cite{liang2022meshmae}, voxels~\cite{hess2023masked}, and Gaussians~\cite{ma2024shapesplat}, allowing the reconstruction of masked regions and facilitating the learning of robust spatial and semantic features. 
In this work, we propose a novel approach for vision-language 3DGS pretraining, as well as a self-supervised learning scheme that operates via Gaussian masking and self-distillation.
\vspace{-2mm}
}
\notoc{\section{\ours Dataset}
We introduce \ourdata ~{--} a carefully curated dataset of 3D Gaussian Splats representing indoor scenes. The main goal was to obtain a dataset to facilitate generalizable 3DGS indoor scene understanding. The dataset contains about seven thousand scenes, including both real-world and synthetic environments. The comprehensive statistics of the introduced dataset are presented in~\cref{tab:dataset}. The \ourdata dataset will be publicly released, with additional details (licenses) available in the supplement.

\subsection{Data Processing}
Multiple measures are applied before, during, and after the 3DGS optimization to guarantee high-quality 3DGS scenes. Starting with the training views, we select scenes with at least 400 frames to ensure sufficient multi-view coverage. 
We remove blurry frames by using the variance of the Laplacian as a sharpness metric.
We use gsplat~\cite{ye2025gsplat} for 3DGS optimization. For scenes with available depth input, we apply depth loss to achieve better geometry modeling. To efficiently compress the 3DGS scene, we employ a Markov Chain Monte Carlo strategy~\cite{kheradmand20253d} and add opacity and scale regularization. Once optimized, we filtered 3DGS scenes based on the PSNR metric before using them as inputs for our pretraining. We refer to the supplement for per-dataset processing details.

\subsection{Data Statistic}
SceneSplat-7K dataset includes various 3D Gaussian Splatting datasets generated from \scannet\cite{dai2017scannet}, \scannetpp\cite{yeshwanth2023scannet++}, \scannetpp v2, \replica\cite{straub2019replica}, Hypersim~\cite{roberts2021hypersim}, 3RScan~\cite{wald2019rio}, \arkit\cite{baruch2021arkitscenes}, and \matterport\cite{chang2017matterport3d}, comprising approximately 9,000 raw scenes. \ourdata contains \textbf{7,916} processed Gaussian splatting scenes, with an average of 1.42 Million 3D Gaussians per scene and \textbf{11.27 Billion Gaussians} in total, from which 4,114 high-quality Gaussian splatting scenes are selected for pretraining. Spanning a total of 4.72 Million RGB frames, SceneSplat-7K achieves high-fidelity appearance and competitive reconstruction quality, having an average PSNR of 29.64 dB, average depth loss of 0.035 m, average SSIM of 0.897, and average LPIPS of 0.212. Constructing this dataset required an equivalent of \textbf{150 days} of computation on an NVIDIA L4 GPU.

\begin{figure*}[!t]

    \centering   \includegraphics[width=\linewidth]{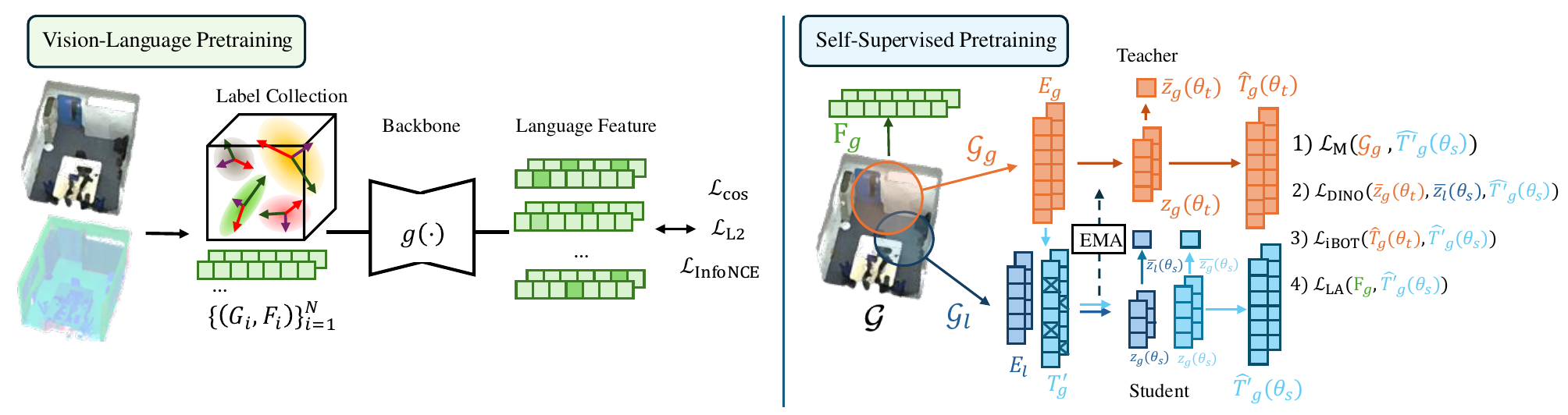}
    \caption{\textbf{\ours Overview}. The SceneSplat-7K dataset enables \textbf{Vision-Language Pretraining} and \textbf{Self-Supervised Pretraining}. For vision-language pretraining, we associate each 3D Gaussian primitive with semantic features based on our label collection process and train a generalizable open-vocabulary learner that predict per-gaussian embeddings. For self-supervised pretraining, we employ Masked Gaussian Modeling to reconstruct masked primitives, Self-Distillation Learning for augmentation-invariant features, and Language-Gaussian Alignment for scenes with collected labels. The former achieves \sota zero-shot segmentation results on ScanNet200~\cite{dai2017scannet}, \scannetpp~\cite{yeshwanth2023scannet++}, and Matterport3D~\cite{chang2017matterport3d} benchmarks and the latter unlocks training on large-scale 3DGS data.
    }

\vspace{-1mm}
\label{fig:pipeline}
\end{figure*}

\section{Methodology}
\label{sec:method}

Building upon the \ourdata dataset, we carry out both vision-language 3DGS pretraining, which enables open-vocabulary scene understanding, and self-supervised pretraining, which regularizes the latent space during 3DGS parameter encoding, as shown in \cref{fig:pipeline}. For vision-language pretraining, we first need to collect primitive-level language labels for 3DGS scenes (\cref{subsec:3dgs_lang_label}). The \ours model then learns to robustly predict vision-language features from the Gaussian parameters and their surroundings (\cref{subsec:lang_pretrain}). Furthermore, for self-supervised pretraining, we employ a multi-objective self-supervised training framework that integrates reconstruction and self-distillation alignment (\cref{subsec:gs_ssl}).

\subsection{3DGS Language Label Collection}
\label{subsec:3dgs_lang_label}

Our language label collection aims to establish 3D-language paired data by associating each 3D Gaussian primitive $G_i$ with a rich semantic feature $F_i \in \mathbb{R}^d$.

Unlike methods that align 3D primitives with text embeddings~\cite{ding2022pla, lee2025mosaic3d} or use visual captioning~\cite{Liu_2024_CVPR, yuan2024osprey}, we directly align Gaussians with the image embedding space of vision-language models (VLM), preserving richer latent semantic information. Our approach also avoids scene-specific compression~\cite{qin2024langsplat,shi2024language} which limits scalability and generalization.

As outlined in Alg.~\ref{alg:label_collection}, we employ SAMv2~\cite{ravi2024sam} for object-level segmentation and SigLIP2~\cite{tschannen2025siglip} for feature extraction. We then use Occam's LGS~\cite{cheng2024occam} to efficiently lift these 2D feature maps to a 3D Gaussian feature field in an optimization-free manner. This results in a comprehensive collection of 3DGS-feature pairs $\{(G_i, F_i)\}_{i=1}^N$ across multiple datasets, providing a solid foundation for our vision-language pretraining.
\begin{algorithm}[t]
    \caption{3DGS Language Label Collection}
    \begin{algorithmic}[1]
    \State \textbf{Input:} Training views $\{I_j\}_{j=1}^M$, 3D Gaussian scene $\{G_i\}_{i=1}^N$, SAMv2, SigLip2
    \State \textbf{Output:} 3D Gaussian-feature pairs $\{(G_i, F_i)\}_{i=1}^N$
    
    \State \textbf{Step 1: 2D Feature Map Generation}
    \For{each training view $I_j$}
        \State $M_{\text{seg}} \gets \text{SAMv2}(I_j)$ \hspace{-10mm} \Comment{Get object-level seg. masks}
        \State $f_g \gets \text{SigLip2}(I_j)$ \Comment{Feature from full frame}
        \State Initialize feature map $F_j$ for view $I_j$
        \For{each segment $s$ in $M_{\text{seg}}$}
            \State $f_l, f_m \gets\text{SigLip2}(\text{crop}(I_j, s))$ \\ \Comment{Local features for crops w/ and w/o background}
            
            \State \textbf{Dynamic Weighting:}
            \State $w_g, w_l, w_m \gets \text{compute\_weights}(f_g, f_l, f_m)$ \\ \Comment{Weights based on context}
            \State $f_s \gets w_g \cdot f_g + w_l \cdot f_l + w_m \cdot f_m$ \\ \Comment{Fuse features}
            
            \State Update feature map $F_j$ with $f_s$ at segment $s$
        \EndFor
    \EndFor
    
    \State \textbf{Step 2: Lifting 2D Features to 3D Gaussian Feature Field}
    \State $\{F_i\}_{i=1}^N \gets \text{Occam's\_LGS}(\{F_j\}_{j=1}^M, \{G_i\}_{i=1}^N)$ \Comment{Lift 2D features to 3D}
    \State $\{F_i\}_{i=1}^N \gets \text{normalize}(\{F_i\}_{i=1}^N)$ \Comment{Normalization}
    
    \State \textbf{Return:} 3D Gaussian-feature pairs $\{(G_i, F_i)\}_{i=1}^N$
    \end{algorithmic}
    \label{alg:label_collection}
\end{algorithm}

\subsection{Vision-Language 3DGS Pretraining} 
\label{subsec:lang_pretrain}

We first adapt the transformer encoder-decoder backbone from~\cite{wu2024point} to efficiently predict high-dimensional per-primitive features corresponding to collected 3DGS language labels.
More specifically, our model $g(\cdot)$, parameterized by $\theta$, maps the input Gaussians to their language features:
\begin{equation}
\hat{F} = g_\theta(\{G_i\}_{i=1}^N) \enspace,
\end{equation}
where $\hat{F} \in \mathbb{R}^{N\times d}$ is the predicted per-gaussian feature.

We apply three training objectives for supervision. The cosine similarity loss minimizes the angular difference between the predicted and ground truth language labels:
\begin{equation}
\mathcal{L}_{\text{cos}} = \frac{1}{|\mathcal{V}|} \sum_{i \in \mathcal{V}} \left(1 - \frac{\hat{F}_i \cdot F_i}{||\hat{F}_i|| \cdot ||F_i||}\right),
\label{eqn:cosine_loss}
\end{equation}
where $\mathcal{V}$ is the set of Gaussians with language feature labels.
To enforce feature similarity in Euclidean space, we use L2 loss:
\begin{equation}
\mathcal{L}_{2} = \frac{1}{|\mathcal{V}|} \sum_{i \in \mathcal{V}} ||\hat{F}_i - F_i||^2 \enspace.
\label{eqn:l2_loss}
\end{equation}

Lastly, we use an aggregated contrastive loss to encourage the separation of class-level features. Instead of contrasting every Gaussian feature individually (which would be computationally prohibitive in large scenes), we apply class-wise \emph{mean pooling}. For each semantic class \(c\) with sufficiently many Gaussians, we randomly split its Gaussians into two disjoint sets \(\mathcal{G}_c^A\) and \(\mathcal{G}_c^B\), and compute the pooled features:
\begin{equation}
\bar{F}_c^A = \text{Pool}(\hat{F}, \mathcal{G}_c^A) \, ,
\quad
\bar{F}_c^B = \text{Pool}(\hat{F}, \mathcal{G}_c^B) \enspace.
\end{equation}
We then apply a bidirectional contrastive loss. First, we normalize \(\bar{F}_c^A\) and \(\bar{F}_c^B\) to unit length. Let
\(Z^A = \bar{F}^A (\bar{F}^B)^\top / \tau\)
and
\(Z^B = \bar{F}^B (\bar{F}^A)^\top / \tau\),
where each row in \(\bar{F}^A\) or \(\bar{F}^B\) corresponds to a different semantic class. The diagonal elements of \(Z^A\) and \(Z^B\) are positive matches. Hence, we compute a cross-entropy loss in both directions:
\begin{align}
\mathcal{L}_{\mathrm{contrast}} 
= \frac{1}{2|C|} \sum_{X \in \{A, B\}} \sum_{i \in C} 
-\log \frac{\exp(Z^X_{i,i})}{\sum_{j \in C} \exp(Z^X_{i,j})} \enspace,
\end{align}
where $C$ is the set of semantic classes with sufficient Gaussians, $\tau$ is a learnable temperature controlling the softness of the distribution.

The total loss is the weighted sum $
\mathcal{L}_{\text{total}} = \lambda_\text{cos} \mathcal{L}_{\text{cos}} + \lambda_\text{L2}\mathcal{L}_{\text{2}} + \lambda_\text{con} \mathcal{L}_{\text{contrast}}
$, where $\lambda_{(\cdot)}$ denotes each weight. In practice, we found that applying the contrastive loss later during the training (\ie, “warm starting” with $\mathcal{L}_{\text{cos}}$ and $\mathcal{L}_{\text{2}}$) helps promote early feature learning while effectively refining class distinctions in later stages.

Through this training, our model learns to predict semantically rich language features for each Gaussian primitive, enabling downstream open-vocabulary scene understanding task without requiring additional finetuning or 2D input.

\subsection{Self Supervised Pretraining}
\label{subsec:gs_ssl}

The proposed GaussianSSL method is presented in \cref{fig:pipeline} (right). It incorporates multiple losses with different objectives into \ours's large-scale pretraining.

\boldparagraph{Masked Gaussian Modeling.}
This part supervises the model to predict masked Gaussian primitives. Given a 3D Gaussian splatting scene represented by $ \mathcal{G} =\{G_i\}_{i=1}^N$, where \( G_i \in \mathbb{R}^{59} \), the training proceeds as follows: 
(1) A subset \( \{G_j\}_{j=1}^{N'} \) is sampled from \( \mathcal{G} \) using dense grid sampling of size \( S \). 
(2) Samples are projected into a latent space with the embedding function \( P \) to obtain tokens $E = P\left(\{G_j\}_{j=1}^{N'}\right) \in \mathbb{R}^{N' \times d_{\text{e}}}$, where \( d_{\text{e}} \) is the embedding dimension. 
(3) The tokens \( E \) are masked with ratio \( r \in [0, 1]\) by replacing \( N' \cdot r \) randomly chosen tokens with a learnable mask token \( t \in \mathbb{R}^{d_{\text{e}}} \) to obtain \( T_{\text{m}} \in \mathbb{R}^{N' \times d_{\text{e}}} \). 
(4) The masked tokens \( T_{\text{m}} \) are processed using the 3D backbone $g_{\theta}(\cdot)$ to obtain $\hat{T_{\text{m}}}=h_{\phi}(f_{\varphi}(T_m))$, where $f_{\varphi}(\cdot)$ is the encoder and $h_{\varphi}(\cdot)$ is the decoder. 
(5) The output tokens $\hat{T_{\text{m}}}$ are mapped to the input Gaussian space with the reconstruction projector $\hat{G}_m = {\Phi}(\hat{T}_m) \in \mathbb{R}^{N' \times F}$. 
(6) Finally, the $\mathcal{L}_2$ reconstruction loss between the predicted masked Gaussians and the original Gaussians is used: 
$\mathcal{L}_{\text{MGM}} = \mathbb{E}_{G_j \sim \mathcal{G}} \left[ \| G_m - \hat{G}_m \|_2^2 \right] $.

\boldparagraph{Self-Distillation Representation Learning.}
Self-distillation~\cite{oquab2023dinov2} learns augmentation-invariant representations by aligning the predictions of a student network $\theta_s$ with an EMA-updated teacher $\theta_t$~\cite{chen2020exploringsimplesiameserepresentation}. For a batch of Gaussian scenes $\{\mathcal{G}_n\}_{n=1}^B$ (global/local views $G_g^b$, $G_l^b$), we extract tokenized bottleneck features $z \in \mathbb{R}^{M \times d_e}$, compute global representations $\bar{z}$ via mean pooling, and align student-teacher outputs via cosine similarity loss $\mathcal{L}_\text{sim}$~\cite{simdino}. Feature diversity is regularized with a coding rate term $\mathcal{L}_\text{cr}$~\cite{simdino,codingrate} resulting in $\mathcal{L}_\text{DINO}=\omega_\text{sim} \mathcal{L}_\text{sim} + \omega_\text{cr}\mathcal{L}_\text{cr} $ with corresponding weights. Inspired by~\cite{zhou2022ibotimagebertpretraining,simdino}, the student network also predicts masked features aligned with the teacher via $\mathcal{L}_\text{iBOT}$, computed using cosine similarity. We propose to mitigate the decoder collapse issues by multi-task reconstruction $\mathcal{L}_\text{MGM}$, as coding rate regularization stabilizes only the hierarchical encoder. Following~\cite{he2022masked,xie2022simmimsimpleframeworkmasked}, the reconstruction is limited to weakly augmented views to avoid degradation.

\boldparagraph{Language-Gaussian Alignment.} As shown in \cref{subsec:lang_pretrain}, the precomputed language feature enables effective knowledge distillation. For scenes with existing language labels, we seek to leverage them to further regulate self-supervised learning. However, the high dimensionality of these language features (dimension $N\times d_L$ ) can substantially increase the computational cost of supervision. To address this, we replace the original features with a compressed representation learned via an autoencoder~\cite{kingma2022autoencodingvariationalbayes}, drastically reducing the memory overhead while preserving semantic information. Similar to $\mathcal{L}_\text{MGM}$, we use $\mathcal{L}_\mathrm{LA}$ following \cref{eqn:cosine_loss,eqn:l2_loss} to train the network to predict low-dimensional language features from unmasked neighbors.

}
\notoc{\section{Experiments}
In this section, we evaluate the performance of vision-language pretraining on open-vocabulary task and examine the impact of large-scale Gaussian self-supervised pretraining on downstream indoor semantic segmentation. We further justify our design choices through ablation studies. The implementation details are provided in the supplement.

\begin{table*}[!t]
    \centering
    \resizebox{\linewidth}{!}{
\begin{tabular}{ll|r|rr|rr|rr}
    \toprule[0.95pt]
    \multirow{2}{*}{Method} 
    & \multirow{2}{*}{Training Source} 
    & \multirow{2}{*}{\shortstack{\#Training \\ Scenes}}
    & \multicolumn{2}{c|}{ScanNet200 (200)}
    & \multicolumn{2}{c|}{Matterport3D (160)}
    & \multicolumn{2}{c}{ScanNet++ (100)} \\
    & & & f-mIoU & f-mAcc & f-mIoU & f-mAcc & f-mIoU & f-mAcc \\
    \midrule[0.6pt]
    OpenScene$^{\dagger}$~\cite{peng2023openscene} 
        & SN & $\times$1 
        & 6.4 & 12.2 
        & 5.7 & 10.7 
        & 8.8 & 14.7 \\ 
    PLA~\cite{ding2022pla} 
        & SN & -- 
        & 1.8 & 3.1 
        & -- & -- 
        & -- & -- \\
    RegionPLC~\cite{yang2024regionplc} 
        & SN & -- 
        & 9.2 & 16.4 
        & 6.2 & 13.3 
        & 11.3 & 20.1 \\
    OV3D~\cite{jiang2024open} 
        & SN & -- 
        & 8.7 & -- 
        & -- & -- 
        & -- & -- \\
    Mosaic3D~\cite{lee2025mosaic3d} 
        & SN & -- 
        & \underline{13.0} & \underline{24.5} 
        & \underline{8.6} & \underline{17.8} 
        & \textbf{16.2} & \textbf{27.1} \\
    \rowcolor{gray!15}\ours (Ours)
        & SN & -- 
        & \textbf{18.9} & \textbf{31.7} 
        & \textbf{10.8} & \textbf{18.7} 
        & \underline{14.7} & \underline{24.7} \\
    \midrule[0.6pt]
    Mosaic3D~\cite{lee2025mosaic3d} 
        & SN, %
        SN++, %
        ARKitS, %
        MP3D, %
        S3D %
        & $\times$24.3 
        & \underline{15.7} & \underline{28.3} 
        & \underline{13.1} & \underline{27.7} 
        & 18.0 & 29.0 \\
    \rowcolor{gray!15}\ours (Ours)
        & SN++ & $\times$0.75 
        & 11.8 & 19.2 
        & 10.6 & 18.6 
        & \underline{26.8} & \underline{45.3} \\
    \rowcolor{gray!15}\ours (Ours)
        & SN, SN++, MP3D & $\times$2.92 
        & \textbf{21.4} & \textbf{38.7}  
        & \textbf{13.8} & \textbf{31.8}   
        & \textbf{28.4} & \textbf{50.0} \\
    \bottomrule[0.95pt]
\end{tabular}
        }
        
    \vspace{-2mm}
    \caption{
        \textbf{Zero-Shot 3D Semantic Segmentation on the Fine-Grained ScanNet++ (100 classes)~\cite{yeshwanth2023scannet++}, Matterport3D (160 classes)~\cite{chang2017matterport3d} and ScanNet200 (200 classes)~\cite{dai2017scannet} Benchmarks.}
        We report the foreground mean IoU (f-mIoU) and foreground mean accuracy (f-mAcc) excluding background classes (wall, floor, ceiling), following~\cite{yang2024regionplc,peng2023openscene,ding2022pla,jia2024sceneverse}.
        $^{\dagger}$ denotes the official checkpoint and the results of the baselines are taken from~\cite{lee2025mosaic3d}.
        Dataset abbreviations SN, SN++, ARKitS, MP3D, and S3D respectively denote ScanNet~\cite{dai2017scannet}, ScanNet++~\cite{yeshwanth2023scannet++},
        ARKitScenes~\cite{baruch2021arkitscenes}, Matterport3D~\cite{chang2017matterport3d} and Structured3D~\cite{zheng2020structured3d}. \ours achieves noticeably better segmentation performance, \ie, a $5.9\%$
        f-mIoU increase on the ScanNet200 benchmark when trained on a single source, and an $11.1\%$ f-mIoU increase on \scannetpp when trained on two sources, while using significantly less training data compared to the concurrent work~\cite{lee2025mosaic3d}.}
    \label{tab:zeroshot_semseg}
\end{table*}

\subsection{Vision-Language Pretraining}
\cref{tab:zeroshot_semseg} reports the zero-shot 3D semantic segmentation results on the fine-grained ScanNet++ (100 classes)~\cite{yeshwanth2023scannet++}, Matterport3D (160 classes)~\cite{chang2017matterport3d} and ScanNet200 (200 classes)~\cite{dai2017scannet} benchmarks, where methods are trained on specified data sources.
When trained on \scannet, \ours achieves \sota results, leading to $5.9\%$ and $2.2\%$ f-mIoU increases on the ScanNet200 and Matterport3D benchmarks. By extending the training sources, we obtain $5.7\%$, $0.7\%$, and $10.4\%$ f-mIoU increases on the ScanNet200, Matterport3D, and \scannetpp benchmarks, respectively, compared to concurrent work~\cite{lee2025mosaic3d}. Notably, \cite{lee2025mosaic3d} uses 8.32$\times$ training scenes to achieve its best results.

\cref{fig:scannetpp_lang_pred_vis} shows the zero-shot segmentation results on evaluation scenes. \ours not only achieves competitive segmentation performance but also correctly annotates the missing semantic labels (\eg, desks shown above). We demonstrate text-based queries on the inference results of the predicted language features in \cref{fig:lang_feat_scene_query}. Our vision-language pretraining enables the effective localization of complex objects within the scene.

\begin{figure}[t]
    \centering
    \begin{tabular}{cc}
    \multicolumn{1}{c}{Zero-Shot Prediction} & \multicolumn{1}{c}{Ground Truth} \\
    \includegraphics[width=0.50\linewidth]{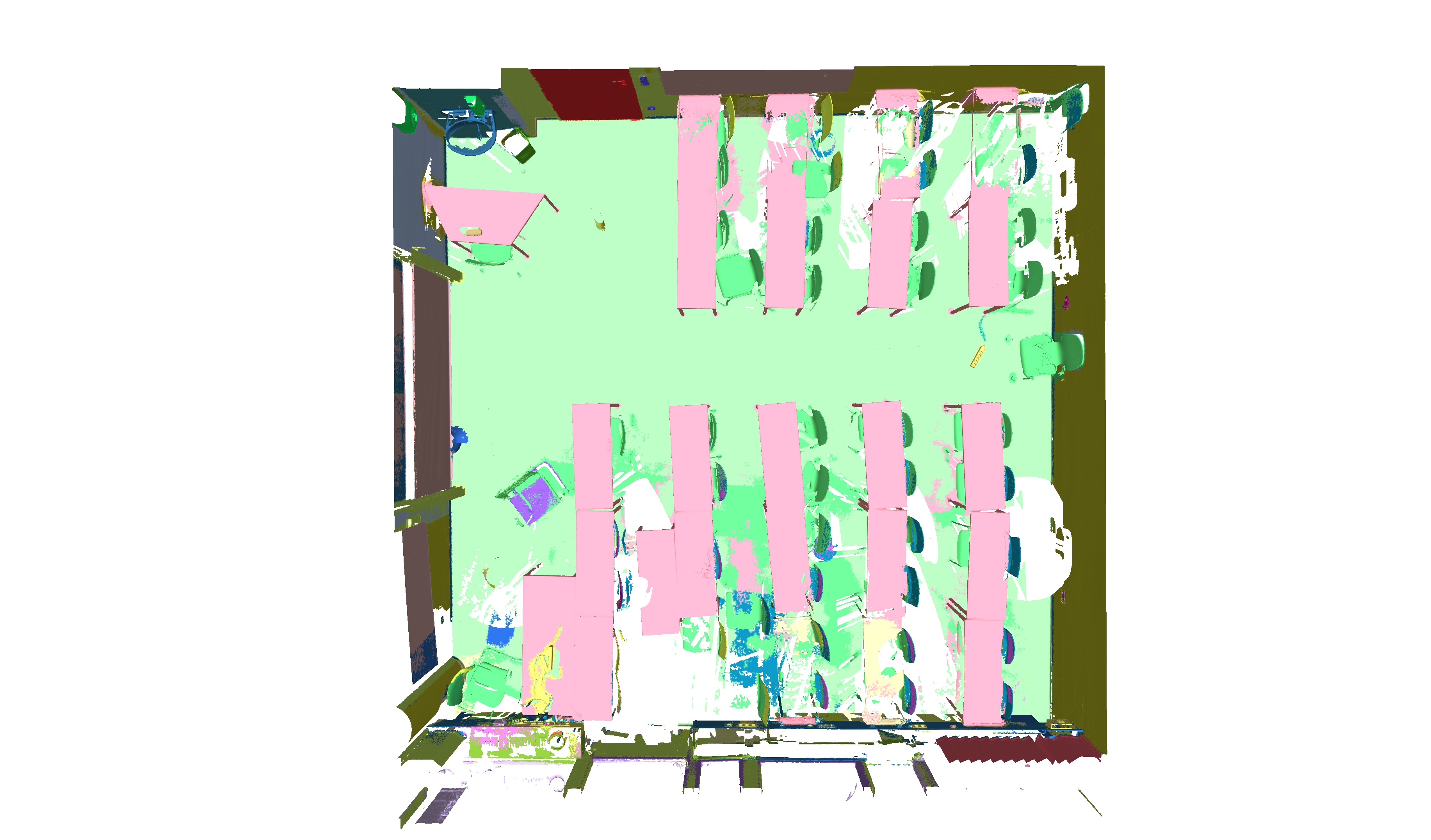} & 
    \includegraphics[width=0.50\linewidth]{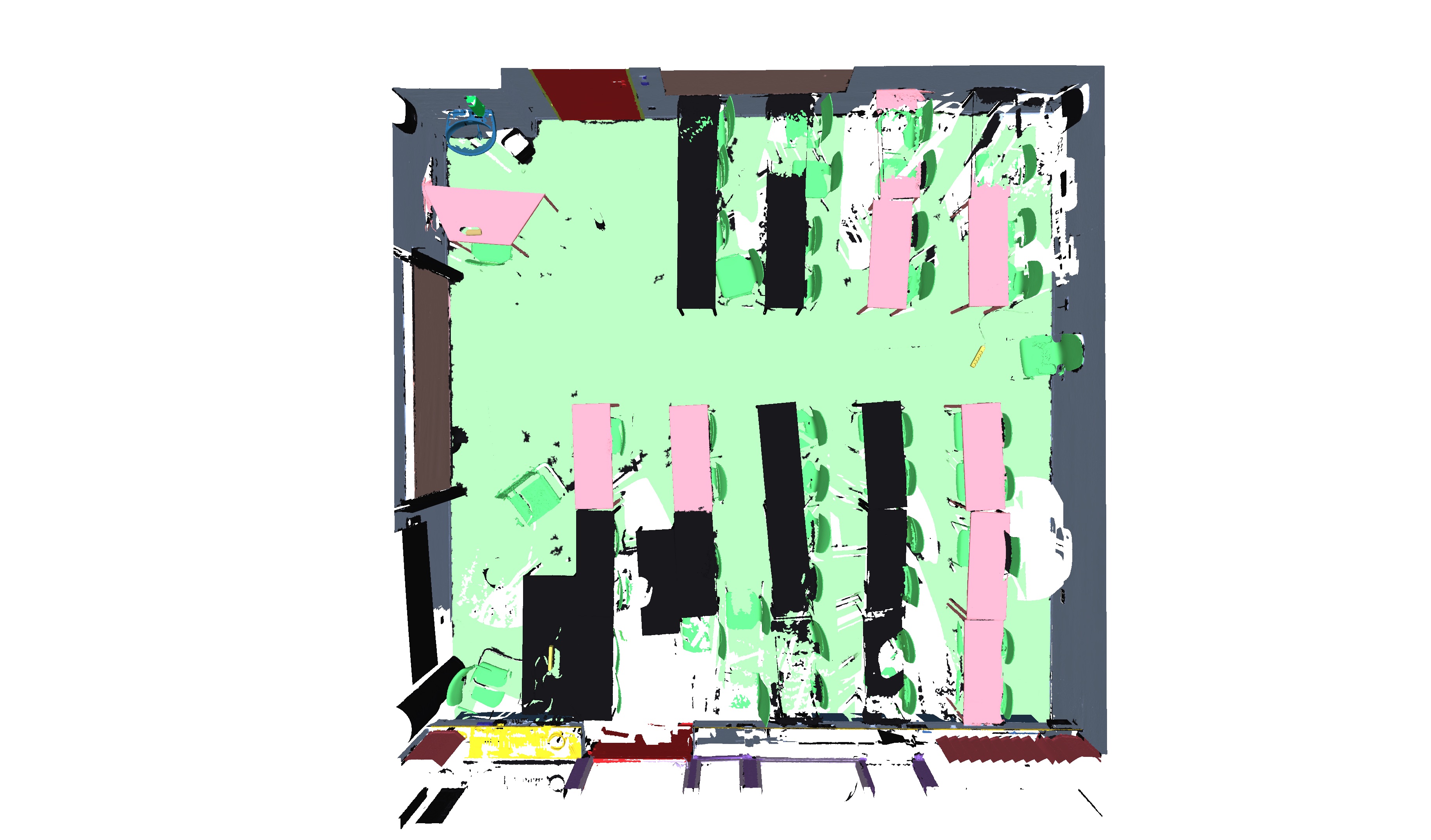} \\
    \includegraphics[width=0.43\linewidth]{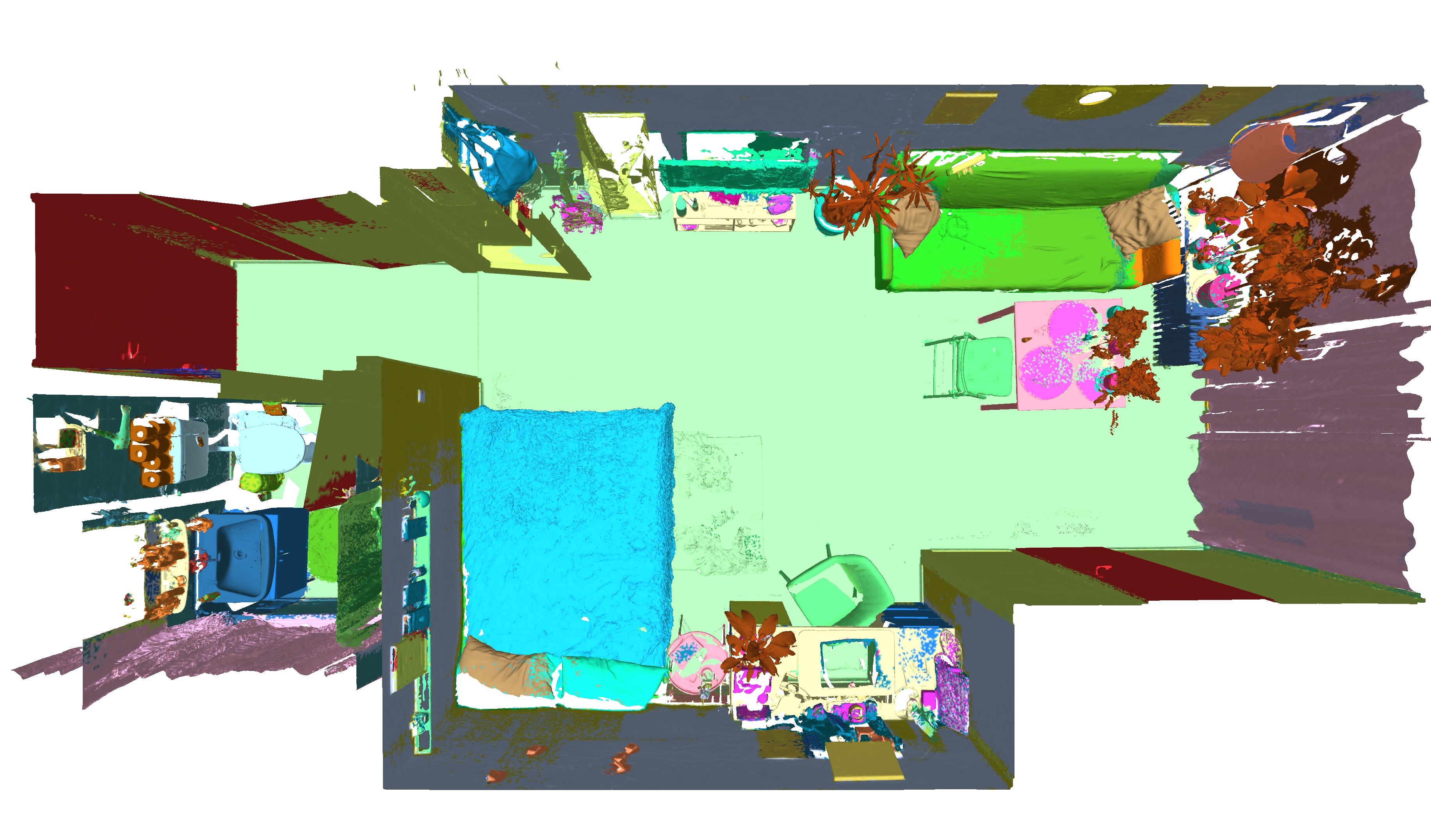} & 
    \includegraphics[width=0.43\linewidth]{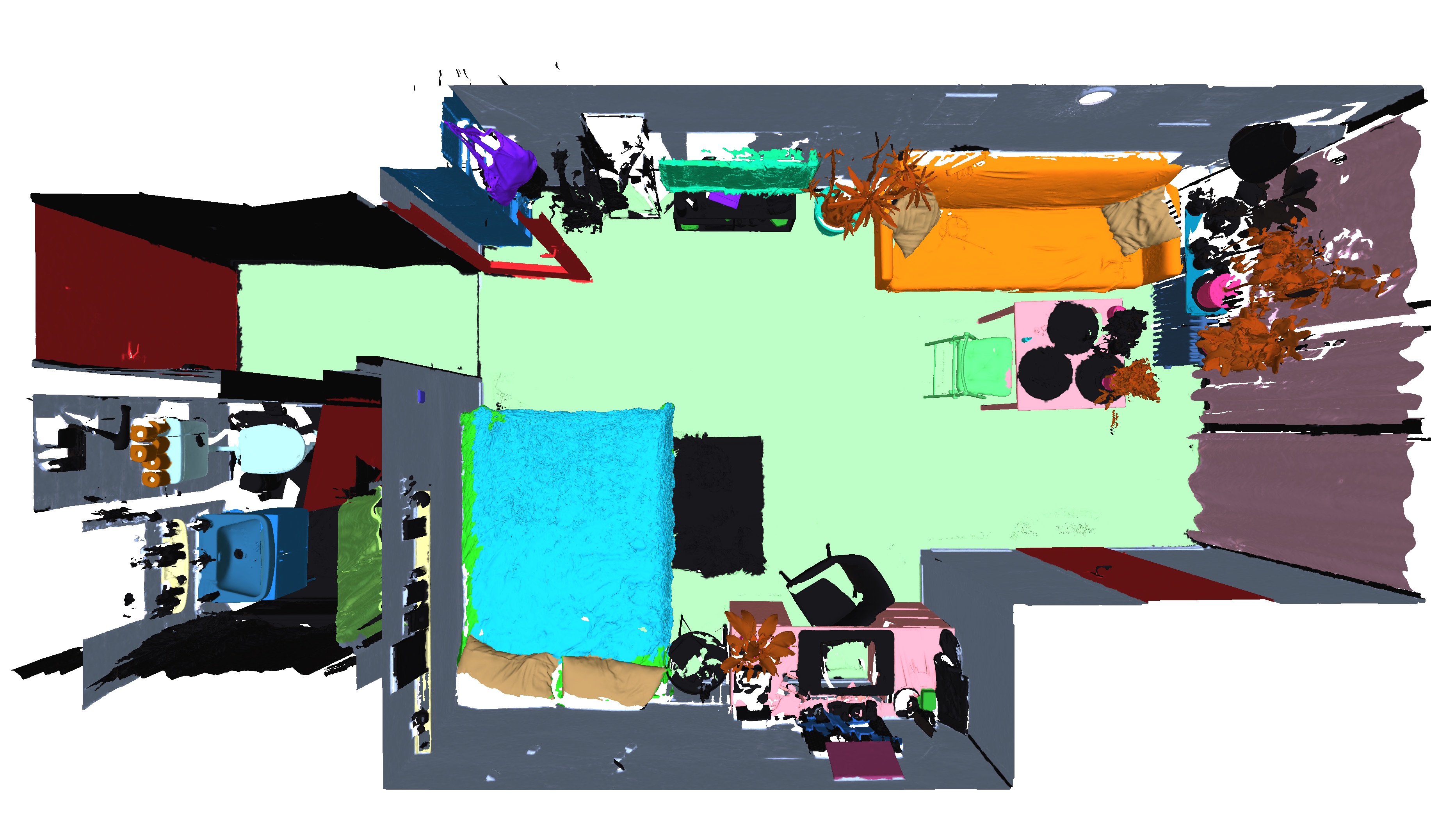} \\
    \end{tabular}
    \vspace{-2mm}
    \caption{
        \textbf{Qualitative Results of Zero-Shot 3D Semantic Segmentation on \scannetpp.} \ours demonstrates competitive zero-shot performance, note how our model correctly annotate the regions lacking ground truth labels, \eg, \textcolor[rgb]{0.9686,0.7137,0.8235}{desks} on the top row. Best viewed zoomed in and in color.
    }
    \label{fig:scannetpp_lang_pred_vis}
\end{figure}

\begin{figure}[t]
\centering
\includegraphics[width=0.48\textwidth]{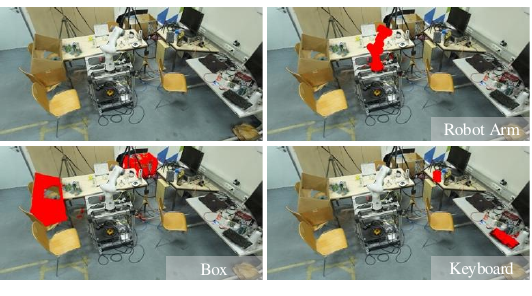}
\vspace{-4mm}
\caption{\textbf{Text-Based 3DGS Scene Query.} Given text queries and \ours inference results for a 3DGS scene, we can effectively localize the corresponding splats (highlighted in \textcolor{red}{red} for queries "Robot Arm", "Box", and "Keyboard").}

\label{fig:lang_feat_scene_query}
\end{figure}

\begin{table}[!t]
    \centering
        \resizebox{\linewidth}{!}{
        \begin{tabular}{l|rr|rr|rr}
            \toprule[0.95pt]
            \multirow{2}{*}{Method} & \multicolumn{2}{c|}{ScanNet20 (20)} & \multicolumn{2}{c|}{ScanNet200 (200)} & \multicolumn{2}{c}{ScanNet++ (100)} \\
             & mIoU & mAcc & mIoU & mAcc & mIoU & mAcc \\
            \midrule[0.6pt]
            No-Pre  & \underline{77.1} & 84.1 & 35.4  & 44.0 & \textbf{42.4} & \textbf{53.3}  \\
            MGM  & 76.7 & 83.5 & \underline{35.5} & \underline{44.5} & 41.7 & 51.9   \\
            +DINO  & 77.0  & \textbf{84.6} & \textbf{35.9} & \textbf{46.1} & \underline{42.0} & \underline{52.4}   \\
            +iBOT  & \textbf{77.2} & \underline{84.2} & 35.2 & 44.3 & 41.1 & 52.0    \\
            +LA  & \textbf{77.2} & \underline{84.2}  & 34.7 & 44.4 & 41.4 & 52.5   \\
            \bottomrule[0.95pt]
        \end{tabular} %
        }
        
        \vspace{-2mm}
        \caption{
            \textbf{GaussianSSL Ablation Experiments.} We adopt the pre-training on the Sceenssplat-7K dataset and report fine-tuning mIOU and mAcc on indoor semantic segmentation tasks. For details of specific losses please refer to \cref{subsec:gs_ssl} and \cref{fig:pipeline}.
        }   
        
    \label{tab:gaussianssl}
\end{table}

\begin{table}[!t]
    \centering
    \footnotesize
    \scalebox{1.0}{
    \begin{tabularx}{\columnwidth}{l|>{\centering\arraybackslash}X>{\centering\arraybackslash}X|>{\centering\arraybackslash}X>{\centering\arraybackslash}X|>{\centering\arraybackslash}X>{\centering\arraybackslash}X}
        \toprule[0.95pt]
            \multirow{2}{*}{Method} & \multicolumn{2}{c|}{ScanNet20 } & \multicolumn{2}{c|}{ScanNet200} & \multicolumn{2}{c}{ScanNet++} \\
             & mIoU & mAcc & mIoU & mAcc & mIoU & mAcc \\
            \midrule[0.6pt]
        PTv1  \cite{zhao2021point}     & 70.6 & --- & 27.8 & --- & ---    & --- \\
        PTv2  \cite{wu2022point}    & 75.4 & --- & 30.2 & --- & ---    & --- \\
        PTv3 \cite{wu2024point}     & 76.4 & 83.5 & 35.0 & 44.2 & \textbf{42.6} & 53.0 \\
        \midrule[0.6pt]
        \rowcolor{gray!15}
        \ours (Ours)     & \textbf{77.2} & \textbf{84.6} & \textbf{35.9} & \textbf{46.1} & 42.4 & \textbf{53.5} \\
        \bottomrule[0.95pt]
    \end{tabularx}
    }
    \vspace{-2mm}
    \caption{
        \textbf{Supervised Semantic Segmentation Experiments.} We report our best results from \cref{tab:gaussianssl} comparing against the state-of-the-art Point Transformer method.
    }
    \label{tab:supervised_semantic}
\end{table}

\subsection{Label-free 3DGS Pretraining}
We conduct extensive pretraining experiments with the proposed GaussianSSL method. 
To assess the efficacy of the pretrained model, we report the segmentation results in \cref{tab:gaussianssl}. Our method achieves a +0.1\% improvement over supervised-only baselines on ScanNet20 and +0.5\% on ScanNet200, while observing a performance drop on ScanNet++ primarily due to pretraining dataset quality variations (\cref{tab:dataset}). Furthermore, compared with our reproduced implementation of PTv3 ~\cite{wu2024point}, we outperform by +0.8\% on ScanNet20 and +0.9\% on ScanNet200 (\cref{tab:supervised_semantic}). More qualitative results are provided in the supplement.

\subsection{Further Statistical Evaluation}
\begin{table}[!t]
    \centering
    \footnotesize
    \begin{tabularx}{\columnwidth}{l|>{\centering\arraybackslash}X>{\centering\arraybackslash}X|>{\centering\arraybackslash}X>{\centering\arraybackslash}X} %
    
    \toprule[0.95pt]
    \multirow{2}{*}{Method} & \multicolumn{2}{c|}{ScanNet200 (200)} & \multicolumn{2}{c}{ScanNet++ (100)} \\
    & f-mIoU & f-mAcc & f-mIoU & f-mAcc \\
    \midrule[0.6pt]
    
    Language Labels & \textbf{22.8} & \textbf{35.9} & 22.6  & \textbf{46.5} \\
    \rowcolor{gray!15}
    \ours & 18.9 & 31.7 & \textbf{26.8} & 45.3 \\
    \bottomrule[0.95pt]
    \end{tabularx}
    
    \vspace{-2mm}
    \caption{\textbf{\ours Inference Results \vs Collected Language Labels on Zero-Shot 3D Semantic Segmentation.} The result on \scannetpp shows the inference performance can be even better than using the collected labels. \ours here is trained on the single dataset respectively.}
    \label{tab:ablation_ours_vs_label}
\end{table}

\boldparagraph{\ours Inference Results \vs Collected Language Labels.}
One may assume that the collected language labels cap the upper bound zero-shot performance since they provide the supervision signal during vision-language pretraining. Interestingly, this is not always the case. \cref{tab:ablation_ours_vs_label} compares the performance of \ours 
inference features with the collected language labels. On \scannetpp, our method outperforms the labels with a $4.2\%$ increase in f-mIoU. 
Although the collected labels are not perfect, large-scale pretraining can filter noise and learn meaningful patterns.

\begin{figure}[t]
    \centering
    \includegraphics[width=0.85\linewidth]{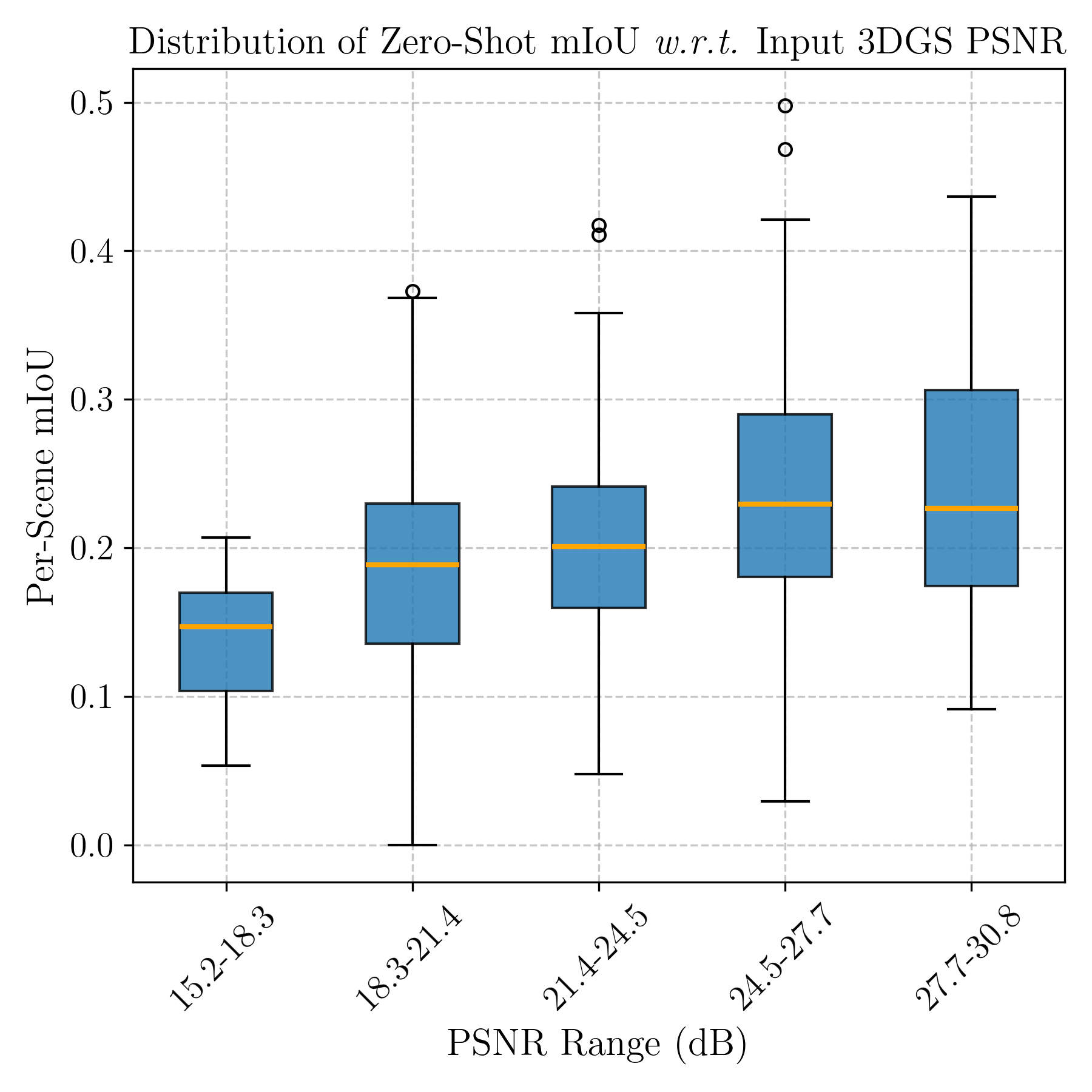}
    \vspace{-2mm}
    \caption{
        \textbf{Distribution of \ours Zero-Shot Semantic Segmentation mIoU \wrt Input 3DGS Scene PSNR.}
        Reported on the Matterport3D test split labeled in 21 semantic classes, the box plot shows a clear positive trend between the input 3DGS scene training PSNR and the resulted mIoU once applied \ours language pretraining for zero-shot semantic segmentation. This encourages the careful curation of the collected 3DGS scene dataset.
    }
    \label{fig:ablation_psnr_iou}
\end{figure}

\boldparagraph{Impact of the Input 3DGS Scene PSNR on Open-Vocabulary Performance.}
Reported on the Matterport3D test split with 370 scenes, \cref{fig:ablation_psnr_iou} indicates a clear positive trend of the input 3DGS scene PSNR of the training views and the resulting zero-shot mIoU performance with \ours model. Low PSNRs usually come out of blurry input images, poor Gaussian centers optimization, and insufficient scene coverage, where the 3DGS parameters cannot resolve the scene well. This trend highlights the importance of data curation for the collected 3DGS scenes.

\begin{figure}[t]
    \centering
    \includegraphics[width=\linewidth]{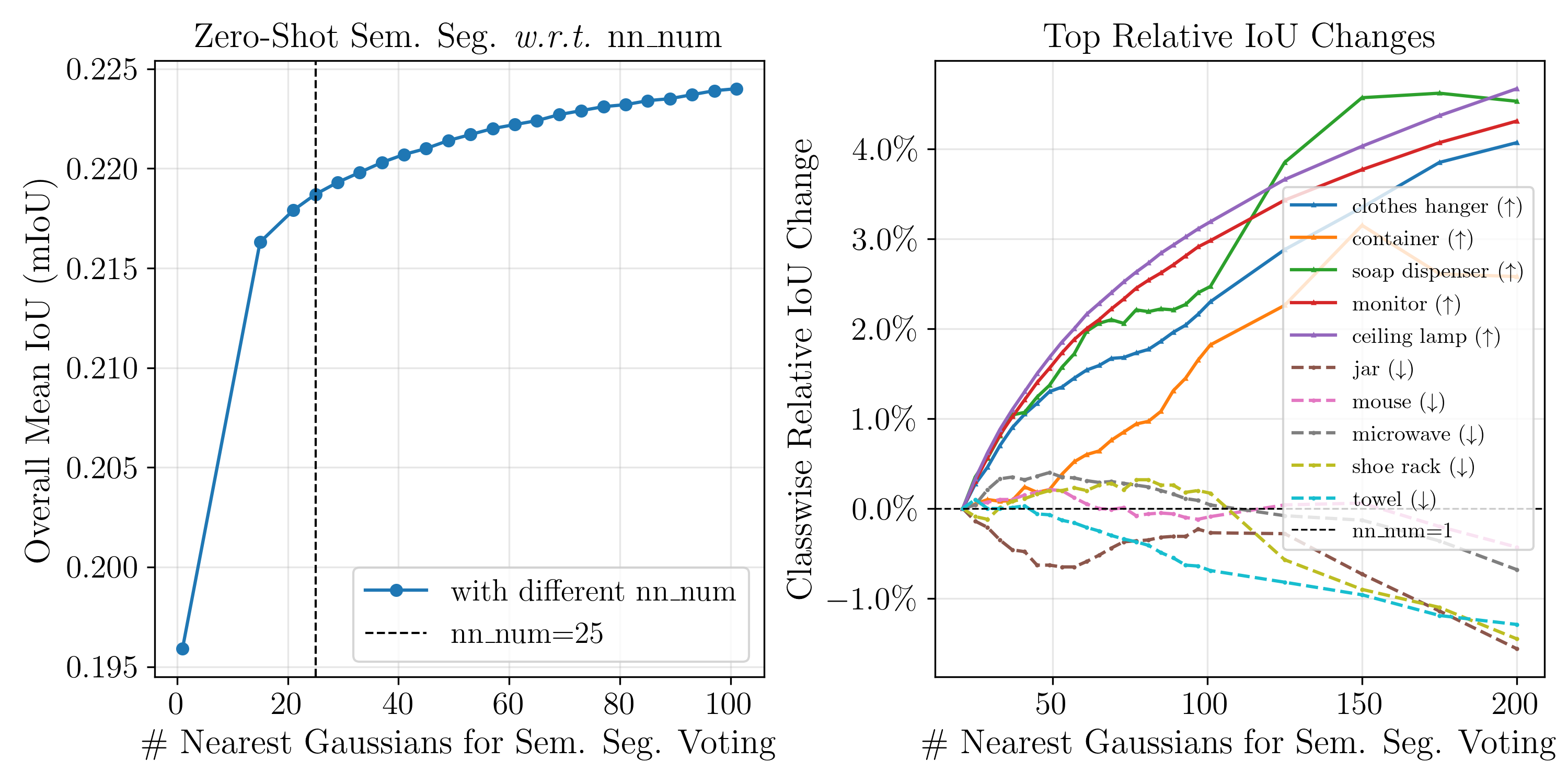}
    \vspace{-2mm}
    \caption{
        \textbf{Overall and Class-Wise IoU Changes \wrt to the Nearest Neighbor Number During Majority Voting.} We evaluate \ours using different nearest 3DGS neighbors for zero-shot task at the point locations on \scannetpp validation split. Overall mIoU increases with different class-wise relative IoU changes.
    }
    \label{fig:ablation_nn_voting}
\end{figure}

\begin{table}[ht!]
    \centering
    \footnotesize
    \begin{tabularx}{\columnwidth}{l|>{\centering\arraybackslash}X>{\centering\arraybackslash}X|>{\centering\arraybackslash}X>{\centering\arraybackslash}X} %
    
    \toprule[0.95pt]
    \multirow{2}{*}{\ours Training Input} & \multicolumn{2}{c|}{ScanNet200 (200)} & \multicolumn{2}{c}{ScanNet++ (100)} \\
    & mIoU & mAcc & mIoU & mAcc \\
    \midrule[0.6pt]
    
    Point Parameters & 17.1 & 27.9   & 23.8 & 40.2 \\
    \rowcolor{gray!15}
    3DGS Parameters & \textbf{18.9} & \textbf{31.7} & \textbf{26.8} & \textbf{45.3}  \\
    \bottomrule[0.95pt]
    \end{tabularx}
    
    \vspace{-1mm}
    \caption{\textbf{Zero-Shot Performance of Using Point Clouds \vs 3DGS for \ours Vision-Language Pretraining.} \ours trained on 3DGS parameters consistently outperforms the variant trained on point cloud properties. The models here are trained on the single dataset respectively.}
    \label{tab:ablation_3dgs_point}
\end{table}

\begin{table}[ht!]
    \centering
    \footnotesize
    \begin{tabularx}{\columnwidth}{l|>{\centering\arraybackslash}X>{\centering\arraybackslash}X|>{\centering\arraybackslash}X>{\centering\arraybackslash}X} %
    
    \toprule[0.95pt]
    \multirow{2}{*}{Contrastive Loss} & \multicolumn{2}{c|}{ScanNet200 (200)} & \multicolumn{2}{c}{ScanNet++ (100)} \\
    & f-mIoU & f-mAcc & f-mIoU & f-mAcc \\
    \midrule[0.6pt]
    
    w/o & 13.7 & 22.5  & 19.6  & 34.4 \\
    always apply & 13.2  & 23.4 & 23.2 & 39.3   \\
    last 75\% epochs & \textbf{15.7} & \textbf{24.0} & \textbf{23.8} & \textbf{40.2} \\
    
    \bottomrule[0.95pt]
    \end{tabularx}
    
    \caption{\textbf{Ablation on Contrastive Loss During Vision-Language Pretraining Using Subsets.} Having a warm-up period and applying the contrastive loss later leads to better performance.}
    
    \label{tab:ablation_contrastive_loss}
\end{table}

\begin{table}[ht!]
    \centering
    \footnotesize
    \begin{tabularx}{\columnwidth}{l|>{\centering\arraybackslash}X|>{\centering\arraybackslash}X} 
    
    \toprule[0.95pt]
    Method & Steps Required & Runtime / Scene  \\
    \midrule[0.6pt]
    
    Occam's LGS & 2D fusion + lifting  &   107 min  \\
    \ours       & single inference & 0.24 min   \\
    
    \bottomrule[0.95pt]
    \end{tabularx}
    
    \caption{\textbf{Runtime.} \ours is significantly faster compared to the current fastest language-embedded 3DGS method.}
    
    \label{tab:ablation_runtime}
\end{table}

\boldparagraph{Nearest Neighbors Voting During Zero-Shot Experiments.}
The centers of the input Gaussians differ from the point locations where the semantic predictions are evaluated; thus, we have to aggregate predictions from neighboring Gaussians. We perform majority voting using the nearest neighboring Gaussians for each evaluation location. \cref{fig:ablation_nn_voting} ablates the number of nearest neighbors on the IoU results using the \scannetpp validation split. We observe the overall trend of the mIoU increase \wrt to the number of nearest neighbors and list the classes with the top relative changes. To balance the performance and inference speed, 25 nearest neighbors are selected for voting. 

\boldparagraph{Effectiveness of Using 3DGS in Vision-Language Pretraining Compared to Point Clouds.}
To further justify the effectiveness of using 3DGS parameters for scene understanding, we apply the same vision-language pretraining on point properties (color and normal). \cref{tab:ablation_3dgs_point} indicates that the model takes point properties as input consistently get outperformed by \ours using 3DGS parameters.

\boldparagraph{Ablation on Contrastive Loss in the Vision-Language Pretraining.}
\cref{tab:ablation_contrastive_loss} ablates the contrastive loss applied during vision-language pretraining, where applying contrastive loss at the late stage of training outperforms other variants.

\boldparagraph{Runtime \vs Per-Scene Language Gaussian Splatting Method.}
Thanks to the feed-forward ability after vision-language pretraining, \ours is shown in \cref{tab:ablation_runtime} to be 445.8$\times$ faster than the \sota language-embedded Gaussian Splatting method, as there is no need for 2D feature extraction and fusion.

\label{sec:experiments}

}
\notoc{\section{Conclusion}
\label{sec:conclusion}
In this work, we introduce \ours, the first large-scale 3D scene understanding model for indoor environments operating directly on 3D Gaussian splats. Powered by \ourdata, a dataset comprising 7,916 scenes, we propose a novel 3D Gaussian splat encoder that generates semantic features in a single pass, enabling open-vocabulary scene recognition without relying on 2D fusion. Through self-supervised techniques, we unlock label-free 3DGS pretraining at the scene level. Our approach achieves state-of-the-art performance in zero-shot semantic segmentation, establishing new benchmarks and laying the foundation for future advancements in open-vocabulary 3D understanding.

}
\notoc{\section{Acknowledgment} 
Yue Li is financially supported by TomTom, the University of Amsterdam and the allowance of Top consortia for Knowledge and Innovation (TKIs) from the Netherlands Ministry of Economic Affairs and Climate Policy. This work used Dutch national e-infrastructure with the support of the SURF Cooperative under grant no. NWO-2024.035. This work was partially supported by INSAIT, Sofia University “St. Kliment Ohridski” (Partially funded by the Ministry of Education and Science of Bulgaria’s support for INSAIT as part of the Bulgarian National Roadmap for Research Infrastructure), 
the MUR PNRR project FAIR (PE00000013), 
the EU Horizon project ELIAS (No. 101120237), 
and the computational resources provided by the Google Cloud Platform (GCP). 
}

\clearpage
\maketitlesupplementary

\setcounter{section}{0}
\setcounter{figure}{0}    
\setcounter{table}{0}   
\setcounter{page}{1}

\definecolor{customblue}{rgb}{0.25, 0.41, 0.88} %
{
\setcounter{tocdepth}{2}   %
\hypersetup{linkcolor=customblue}
\tableofcontents
}

\renewcommand{\thetable}{\Alph{table}}
\renewcommand{\thefigure}{\Alph{figure}}
\renewcommand{\thesection}{\Alph{section}}

\section{Implementation Details}

\subsection{Language Label Collection}
\label{subsec:supp_lang_label}

Many existing methods align 3D primitives with text embeddings from vision-language models (VLM)~\cite{ding2022pla, lee2025mosaic3d}. While effective for basic understanding, this approach inherently limits the information captured, as textual descriptions typically only convey categorical and spatial properties, lacking fine-grained visual details and relationships. Methods employing visual captioning models~\cite{Liu_2024_CVPR, yuan2024osprey} face similar challenges as they struggle to describe all aspects of the target scene content. Even detailed captions inevitably result in information loss during the text embedding alignment process. In contrast, we directly align Gaussians with the image embedding space of VLM, thereby preserving richer latent semantic information.

\boldparagraph{Dynamic Weighting Mechanism.}
Unlike previous approaches that use a single tight crop around each segment, we adapt the three-crop strategy~\cite{werby2024hierarchical} with dynamic weighting to capture the context. For each segment identified by SAMv2, we extract three distinct features: (1) global feature $f_g$ from the entire RGB frame, capturing the full scene context; (2) local feature $f_l$ from the image crop with background; (3) masked feature $f_m$ from the image crop without background, focusing solely on the object. During the process, SAMv2/sam2-hiera-large and SigLIP2/siglip2-base-patch16-512 models are used. 

Our dynamic weighting mechanism combines the three extracted features through the following equations. First, we calculate the cosine similarity between features with and without background and use it to create a fused local feature:
\begin{align}
r_{lm} &= \text{sim}(f_l, f_m)\\
F_l &= r_{lm} \cdot f_m + (1 - r_{lm}) \cdot f_l,\\
F_l &= \text{normalize}(F_l)
\end{align}

We then compute the similarity between this fused local feature and the global feature to determine the final weights:
\begin{align}
\phi_{lG} &= \text{sim}(F_l, f_g), \quad w_i = \text{softmax}(\phi_{lG})\\
w_g &= w_i, \quad w_m = (1 - w_i) \cdot r_{lm}, \\
w_l &= (1 - w_i) \cdot (1 - r_{lm})
\end{align}

The final fused feature is computed as a weighted combination and normalized as:
\begin{align}
f_s &= w_g \cdot f_g + w_l \cdot f_l + w_m \cdot f_m, \quad f_s = \text{normalize}(f_s)
\end{align}

This mechanism adaptively balances the global context influence via $w_g$, local context with background via $w_l$, and object-specific features via $w_m$. These three features are dynamically combined to create a representation that adapts to the segment’s relationship with its context. For objects that are highly integrated with their surroundings (\eg, a keyboard in front of a monitor), background-inclusive features receive a higher weight. For isolated objects (\eg, a coffee mug), background-excluded features dominate.

\subsection{Vision-Language Pretraining}
Our vision-language 3DGS pretraining model is built on a transformer encoder-decoder architecture adapted from~\cite{wu2024point}. As detailed in \cref{tab:language_ptv3}, the encoder consists of 4 stages with depths [2,2,2,6], channels [32,64,128,256], and heads [2,4,8,16], while the decoder employs 3 stages with depths [2,2,2], channels [768,512,256], and 16 attention heads each. We process 3D Gaussian primitives with all parameters (center, color, opacity, quaternion, and scale). The model is trained with the AdamW optimizer (initial LR = 0.006, weight decay = 0.05) using a OneCycle scheduler with cosine annealing. Our loss weights are set to $\lambda_\text{cos}=1.0$, $\lambda_\text{L2}=1.0$, and $\lambda_\text{con}=0.02$, with contrastive loss (temperature $\tau=0.2$) activated only in the later $75\%$ of training. During training, we employ extensive data augmentation, including random rotations, scaling, flipping, jittering, and elastic distortions (see \cref{tab:ptv3_data_aug}). We train for 800 data epochs using 4 NVIDIA H100 GPUs.

\subsection{Masked Gaussian Modeling}
For all self-supervised pretraining experiments on the full dataset, we employ 4 NVIDIA H100 GPUs (94GB each) and the AdamW optimizer, with learning rates of \(1 \times 10^{-3}\) for the embedding layer and \(1 \times 10^{-4}\) for the attention blocks, coupled with a weight decay of \(1 \times 10^{-3}\). We use mixed-precision training (FP16) to accelerate convergence and reduce memory overhead. In total, we train for 300k steps. We use an $\mathcal{L}_2$ loss to enforce consistency between predicted and ground truth Gaussian parameters, focusing only on masked regions. For parameter reconstruction, we design a three-layer MLP projector for each Gaussian attribute, incorporating output activations that adhere to physical constraints: color uses a $\tanh$ activation (bounded in $[-1, 1]$), opacity and scale use a sigmoid activation (bounded in $[0, 1]$) because in most indoor scenes, most Gaussians have small scales, rotation, and normals apply $\tanh$ followed by \( \ell_2 \)-normalization to ensure valid quaternions and normals.

\subsection{Self-Distillation Representation Learning}
In self-distilled representation learning, we adopt the framework of \cite{simdino}, combining DINO loss and coding rate regularization on pooled encoder features. We configure a batch size of 24 and set the grid sampling resolution to 0.02 for partitioning Gaussian scenes. After grid sampling, the scenes are randomly partitioned into base views \(\mathcal{G}_b\), which are then used to generate global and local crops.  For global crops, we randomly select a Gaussian splat from \(\mathcal{G}_b\) and sample its \(K\) neighbors, where \(K \sim [0.4, 1.0] \times 256,000\). For local crops, we sample \(K \sim [0.1, 0.4] \times 102,400\). Each batch includes $N_g=2$ global views and $N_l=3$ local views to balance the scene coverage and granularity. For the DINO loss \( \mathcal{L}_{\text{DINO}} \), we extract tokenized features from the student encoder \( f_{\theta}(\cdot) \) and teacher encoder \( f_{\phi}(\cdot) \), where \( \phi \) is updated via exponential moving average (EMA). A global representation \( \bar{z} \) is obtained by mean pooling over spatial tokens. This representation is projected into a latent space using a 3-layer MLP $P_\text{DINO}$ (dimensions: 2048 → 2048 → 256) followed by \( \ell_2 \)-normalization. Then, we calculate the similarity loss:
\begin{equation}
    \mathcal{L}_{\text{sim}} = \frac{1}{N_g \times N_l} \sum_{i=1}^{N_l} \sum_{j=1}^{N_g} \frac{P_\text{DINO}(\bar{z_{l}}^{(i)}) \cdot P_\text{DINO}(\bar{z_g}^{(j)})}{\|P_\text{DINO}(\bar{z_{l}}^{(i)})\| \| P_\text{DINO}(\bar{z_g}^{(j)})\|}
\end{equation}

For regularization, we use the negative of the coding rate:

\begin{equation}
    R_{\epsilon}(\Gamma) := \frac{1}{2}\text{logdet}\left(\mathbf{I} + \frac{d}{\epsilon^{2}} \Gamma   \right),
\end{equation}
\( R_{\epsilon} \) approximates the rate-distortion function of a Gaussian random variable with covariance \( \Gamma \), becoming exact as \( \epsilon \to 0 \). More specifically, it quantifies the spread of covariance, even when the underlying variables are not strictly Gaussian. 

For the iBOT loss, we mask 50\% of the global views with masking ratios sampled uniformly from \( r \sim [0.2, 0.7] \). The masked tokens are replaced by a learnable token (\( T_{\text{mask}} \)). Note here the masked global view embedding feature goes through the student network and outputs $\hat{T'}(\theta_s)  \in \mathbb{R}^{N' \times d_{\text{r}}} $, where $d_{\text{r}}$ denotes the representation dimension, while the unmasked global view embedding is forwarded to the teacher network and outputs $\hat{T'}(\theta_t) \in \mathbb{R}^{N' \times d_{\text{r}}} $. We detach the teacher's output and calculate the simple iBOT loss using 3-layer MLP $P_\text{iBOT}(\cdot)$ (dimensions: 256 → 256 → 32) as the projector:
\begin{equation} 
    \mathcal{L}_{\text{iBOT}} =  \frac{1}{|M|} \sum_{i \in M}  \underbrace{
        \frac{ P_\text{iBOT}(\hat{T'}(\theta_s)^{(i)}) \cdot P_\text{iBOT}(\hat{T'}(\theta_t)^{(i)})}
        {\left\|  P_\text{iBOT}(\hat{T'}(\theta_s)^{(i)} )\right\| \left\|  P_\text{iBOT}( \hat{T'}(\theta_t)^{(i)})\right\|}
    }_{\text{Pairwise Cosine Similarity}}
\label{equ:ibot}
\end{equation}

Only the masked regions $M$ contribute to the iBOT loss calculation. Following \cite{simdino}, we simplify the original iBOT implementation by removing the centering operation and online tokenizer and replacing them with a pairwise cosine-similarity loss for feature alignment. When MGM loss is enabled, for one global view, we will employ less aggressive augmentation, and this view will contribute to both iBOT loss and MGM loss.
\subsection{Autoencoder for Feature Compression}
\label{supp:sub:ae}
Building on the idea from \cite{qin2024langsplat}, we train a scene-specific language autoencoder on precomputed vision-language embeddings. This model compresses high-dimensional SigLIP features into a low-dimensional latent space, enabling the efficient storage of language features within Gaussian representations. The encoder (by default 5-layer architecture: [384, 192, 96, 48, 16]) generates compact latent codes, while the symmetric decoder (by default 5-layer architecture: [48, 96, 192, 384, 768]) reconstructs the original high-dimensional embeddings. This design reduces memory overhead while preserving semantic fidelity. Unlike \cite{qin2024langsplat}, our model is trained on 3D Gaussian features rather than 2D image-level data. For detailed ablations on the autoencoder architecture and training paradigm, see \cref{tab:autoencoder_abaltion}.

\subsection{Gaussian Language Alignment.}
We follow the same loss functions in \cref{eqn:cosine_loss,eqn:l2_loss} for feature alignment. When combining MGM with language alignment, we adhere to the MGM process by masking the input tokens and applying compressed language alignment loss only to the masked regions. Given the comparable magnitudes of all loss terms, we assign equal weights (\(\omega_{\text{MGM}} = \omega_{\text{DINO}} = \omega_{\text{iBOT}} = \omega_{\text{LA}} = 1.0\)) to maintain the balance of different objectives during training.

\subsection{Model Architecture}
\label{supp:model_design}

\begin{table}
    \small
    \begin{tabularx}{\linewidth}{l X}
    \toprule
    Config &Value \\\midrule
    embedding depth &2 \\
    embedding channels &32 \\
    encoder depth &[2, 2, 2, 6] \\
    encoder channels &[32, 64, 128, 256] \\
    encoder num heads &[2, 4, 8, 16] \\
    encoder patch size &[1024, 1024, 1024, 1024] \\
    decoder depth &[2, 2, 2] \\
    decoder channels &[768, 512, 256] \\
    decoder num heads &[16, 16, 16] \\
    decoder patch size &[1024, 1024, 1024] \\
    down stride &[$\times$2, $\times$2, $\times$2, $\times$2] \\
    mlp ratio &4 \\
    qkv bias &True \\
    drop path &0.3 \\
    \bottomrule
    \end{tabularx}
    \caption{\textbf{Model Configs for Vision-Language Pretraining.} }
    \label{tab:language_ptv3}
\end{table}

\begin{table}
\small
\begin{tabularx}{\linewidth}{l X}
\toprule[0.95pt]
Config & Value \\\midrule[0.6pt]
embedding depth &2 \\
embedding channels &32 \\
encoder depth &[2, 2, 2, 6, 2] \\
encoder channels &[32, 64, 128, 256, 512] \\
encoder num heads &[2, 4, 8, 16, 32] \\
encoder patch size &[1024, 1024, 1024, 1024] \\
decoder depth &[2, 2, 2, 2] \\
decoder channels &[64, 64, 128, 256] \\
decoder num heads &[4 4, 8, 16] \\
decoder patch size &[1024, 1024, 1024] \\
down stride &[$\times$2, $\times$2, $\times$2, $\times$2] \\
mlp ratio &4 \\
qkv bias &True \\
drop path &0.3 \\
\bottomrule[0.95pt]
\end{tabularx}
\caption{\textbf{Model Configs for GaussianSSL Pretraining and Downstream Semantic Segmentation.} }
\label{tab:semantic_ptv3}
\end{table}

\begin{table*}
    \centering
    \begin{tabular}{llcc}
    \toprule[0.95pt]
    Augmentations &Parameters &Global View & Local View \\
    \midrule[0.6pt]
    \multicolumn{2}{c}{Base Transform} & \checkmark  &  \checkmark  \\
    \midrule[0.6pt]
    random rotate &axis: z, angle: [-1, 1], p: 0.5 &  & \\
    &axis: x, angle: [-1 / 64, 1 / 64], p: 0.5 &  &  \\
    &axis: y, angle: [-1 / 64, 1 / 64], p: 0.5 &  &  \\
    random scale &scale: [0.9, 1.1] &  &  \\
    random flip &p: 0.5 &  &  \\
    random jitter &sigma: 0.005, clip: 0.02 &  &  \\
    elastic distort & params: [[0.9, 0.1]] &  &  \\
    grid sampling & grid size 0.02 &  & \\
    
    \midrule[0.6pt]
    \multicolumn{2}{c}{Global Base Transform} & \checkmark  &  -  \\
    \midrule[0.6pt]
    random flip &p: 0.5 &  &  \\
    random crop & ratio: (0.4, 1.0) max: 256000 &  & \\ 
    \midrule[0.6pt]
    \multicolumn{2}{c}{Global Transform 0} & \checkmark  &  -  \\
    \midrule[0.6pt]
    random color jitter &b:0,4, c:0.4, s:0.2 hue:0.1 p:0.8 & & \\
    random grayscale &p: 0.2 &  &  \\
    random Gaussian blur & p: 1.0 & & \\
    \midrule[0.6pt]
    \multicolumn{2}{c}{Global Transform 1} & \checkmark  &  -  \\
    \midrule[0.6pt]
    random dropout &dropout ratio: 0.2, p: 0.2 & & \\
    random color jitter &b:0,4, c:0.4, s:0.2 hue:0.1 p:0.8 & & \\
    random grayscale &p: 0.2 &  &  \\
    random Gaussian blur & p: 0.2 & & \\
    \midrule[0.6pt]
    \multicolumn{2}{c}{Local Base Transform} & - &  \checkmark   \\
    \midrule[0.6pt]
    elastic distort & params: [[0.2, 0.4], [0.8, 1.6]] & & \\
    random flip &p: 0.5 &  &  \\
    random crop & ratio: (0.1, 0.4) max: 102400 &  & \\ 
    \midrule[0.6pt]
    \multicolumn{2}{c}{Local Transform} & - &  \checkmark   \\
    \midrule[0.6pt]
    random dropout &dropout ratio: 0.2, p: 0.2 & & \\
    random color jitter &b:0,4, c:0.4, s:0.2 hue:0.1 p:0.8 & & \\
    random grayscale &p: 0.2 &  &  \\
    random Gaussian blur & p: 0.5 & & \\
    \bottomrule[0.95pt]
    \end{tabular}
    \caption{\textbf{Data Augmentations.} Following \cite{oquab2023dinov2}, we design the data augmentations for 3D scenes for global and local views.}
    \label{tab:ptv3_data_aug}
\end{table*}

The network design choices are described in detail in \cref{tab:language_ptv3,tab:semantic_ptv3}. For the 3D backbone, we adopt the state-of-the-art Point Transformer~\cite{wu2024point}, enhanced with Flash Attention, to substantially improve computational efficiency. To optimize feature embedding for scene-level Gaussians, we use sparse convolutions, which preserve geometric details while minimizing the memory overhead.

\section{Further Experimental Results}
\label{supp:more_results}

\subsection{\ours Zero-shot Segmentation}
\begin{figure}[ht]
    \centering
    \includegraphics[width=0.95\linewidth]{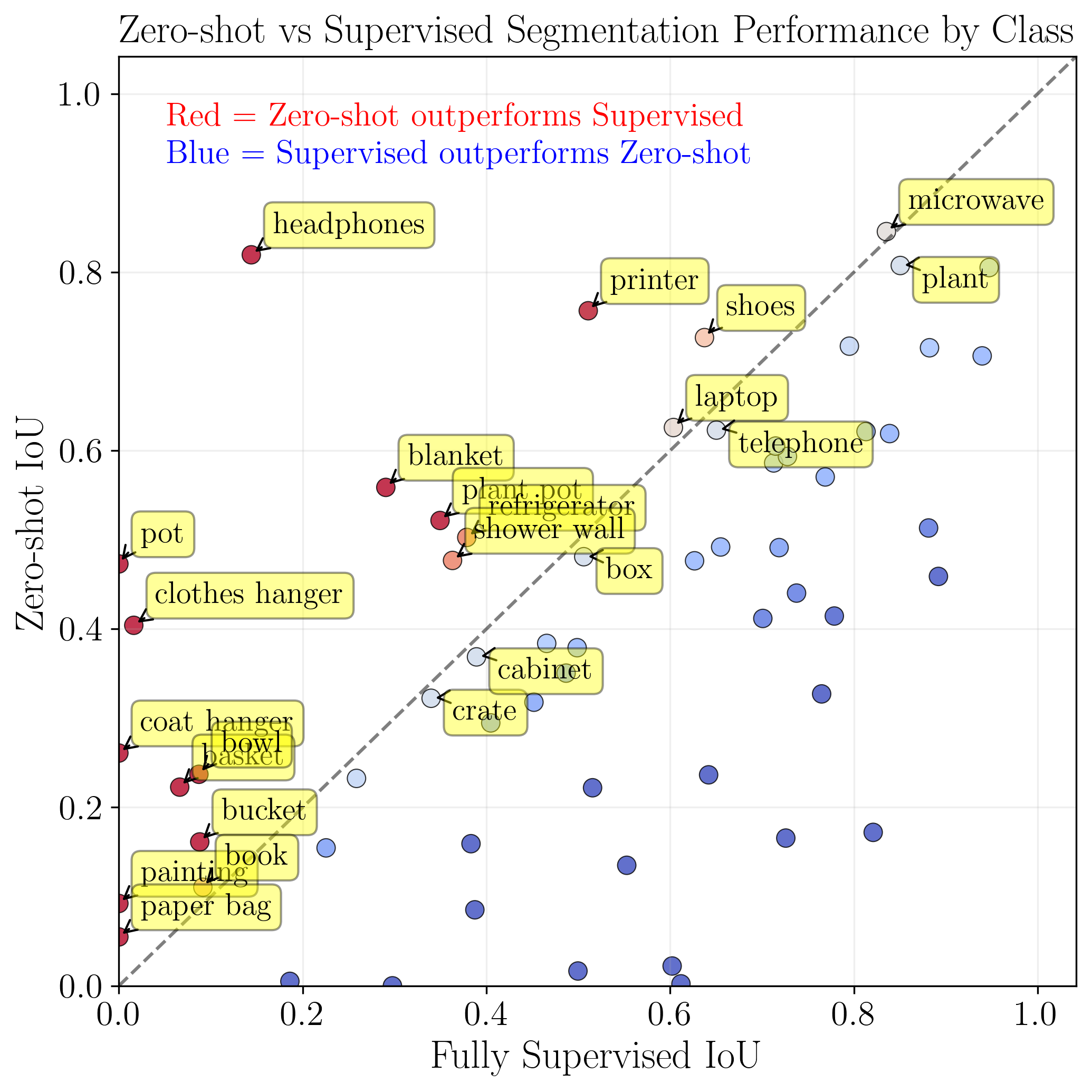}
    \vspace{-2mm}
    \caption{
        \textbf{Comparison of \ours Zero-Shot Versus Fully Supervised Segmentation Across Object Classes.} Notably, our zero-shot segmentation after vision language pretraining demonstrates better performance for 18 object classes, predominantly small objects such as headphone, printer, pot, and clothes hanger.   
    }
    \label{fig:zeroshot_vs_sup}
\end{figure}

\boldparagraph{Zero-shot \vs Fully Supervised.}
\cref{fig:zeroshot_vs_sup} compares the class-wise zero-shot 3D segmentation results with those of the current \sota supervised method~\cite{wu2024point}. Points above the diagonal line (colored red) represent classes where zero-shot segmentation outperforms fully supervised approaches, whereas points below (colored blue) indicate the opposite. Our zero-shot segmentation achieves superior performance for 18 object classes in the \scannetpp benchmark, predominantly small objects such as headphones, printers, pots, and clothes hangers. These results demonstrate the robust prior knowledge acquired by our model through vision-language pretraining.

\begin{figure*}[t]
\centering

\includegraphics[width=0.90\textwidth]{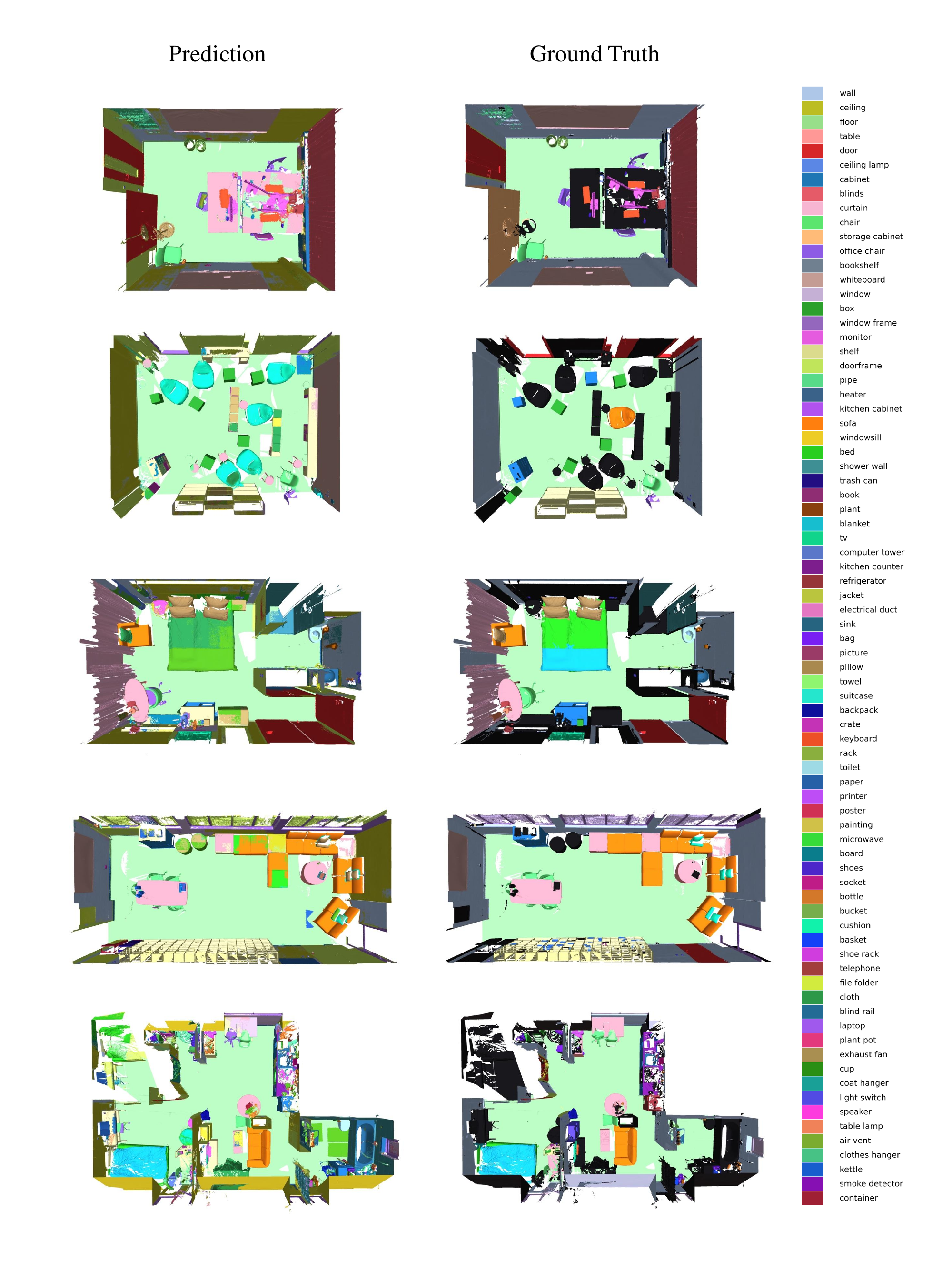}

\caption{\textbf{Qualitative Zero-shot Semantic Segmentation Results on \scannetpp Validation Scenes.} \ours effectively segments the scenes and helps annotate regions with missing labels in the ground truth.}
\label{fig:land_pred_vis_supp}
\end{figure*}

\boldparagraph{More Qualitative Zero-shot Segmentation Results.} \cref{fig:land_pred_vis_supp} presents more qualitative zero-shot semantic segmentation results on \scannetpp validation scenes. \ours effectively segments the scenes and helps annotate regions with missing labels in the ground truth.

\begin{figure}[ht]
    \centering
    \begin{tabular}{cc}
    \includegraphics[width=0.45\linewidth]{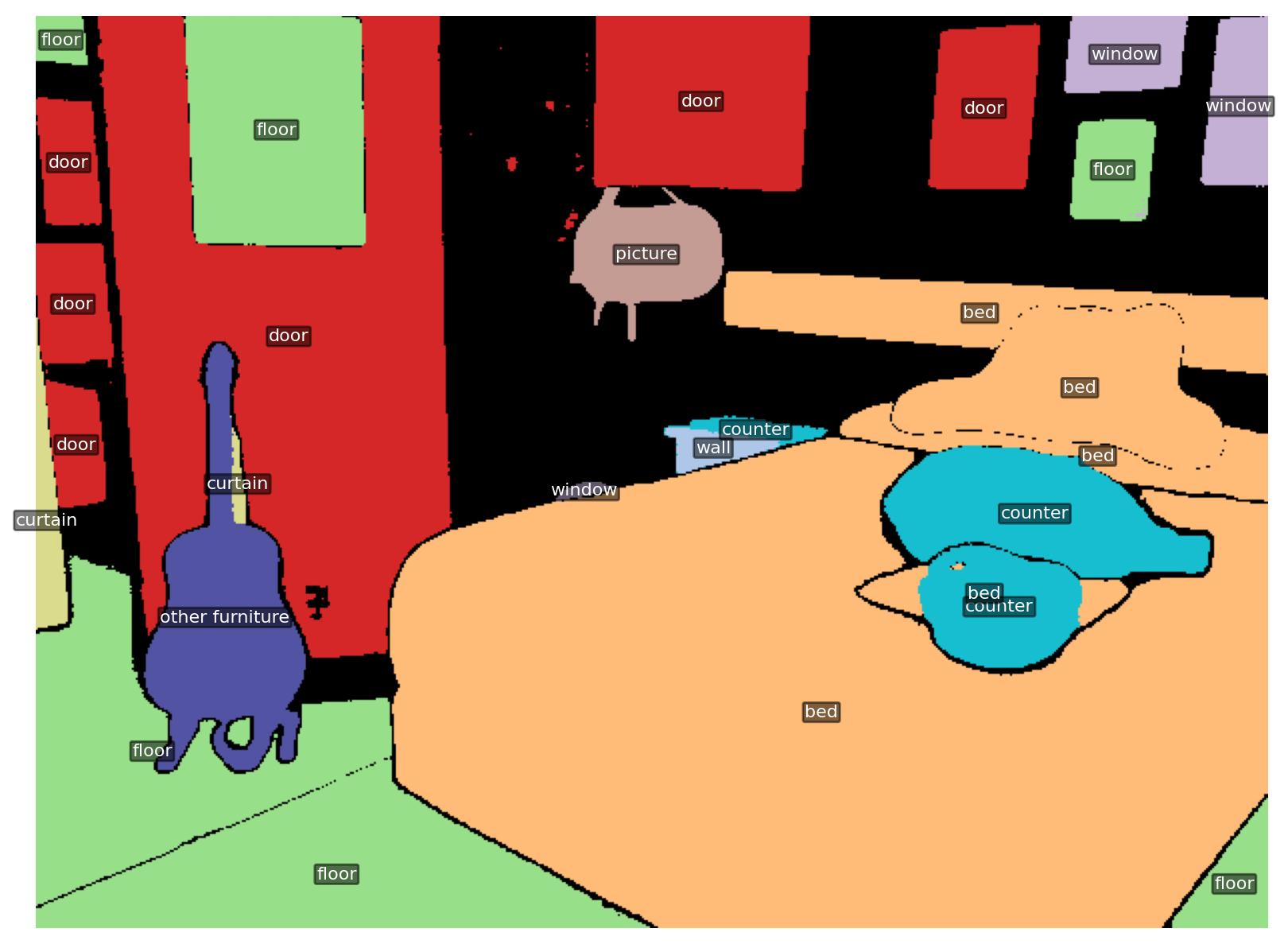} & 
    \includegraphics[width=0.45\linewidth]{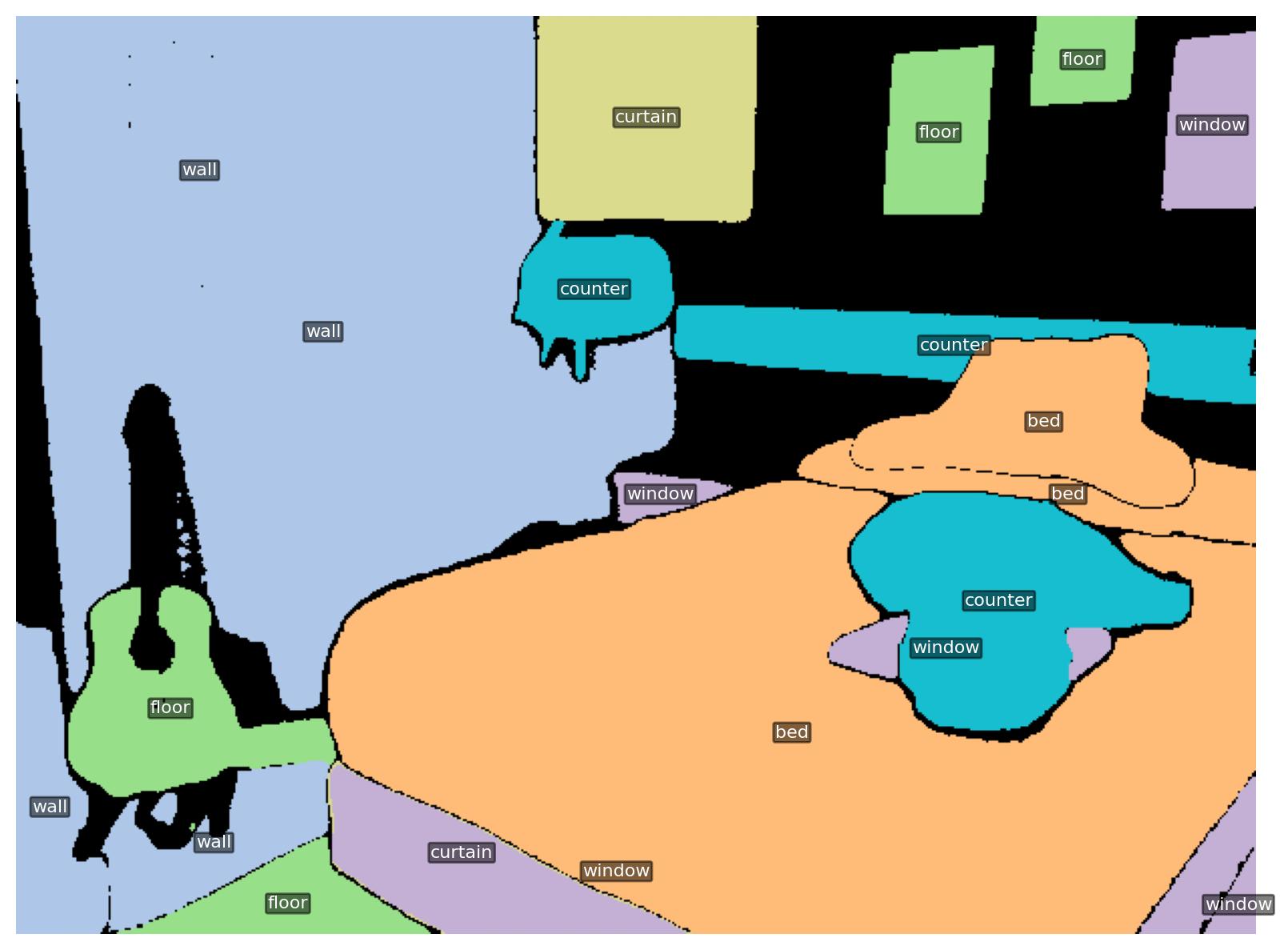} \\
    \end{tabular}
    \vspace{-1mm}
    \caption{
        \textbf{Consistency Issue During 2D Vision-Language Feature Map Collection on \scannet.} Erroneous 2D feature maps are collected, as shown by the mislabeled regions in the neighboring figures. The root cause is that the SAMv2+SigLip2 process we use does not guarantee temporal consistency.
    }
    \label{fig:scannet20_label_issue}
\end{figure}

\boldparagraph{Consistency Issue in Label Collection.} 
\cref{tab:scannet20_semseg} presents our zero-shot segmentation results on the ScanNet20 benchmark. We observe that the performance on this particular benchmark is significantly lower compared to the \sota results we achieve on the other three benchmarks. Through a detailed analysis, we identify that the issue stems from inconsistencies in the 3DGS language label collection process. Specifically, the SAMv2+SigLIP2 pipeline that we employ does not guarantee temporal consistency, especially for large background objects such as walls and floors, as illustrated in \cref{fig:scannet20_label_issue}. This inconsistency results in corrupted feature fields for Gaussians associated with these regions, subsequently leading to reduced zero-shot segmentation performance. Addressing this temporal consistency issue remains an important direction for future research.

\begin{table}[ht]
    \centering
    \footnotesize
    \begin{tabularx}{\columnwidth}{l|>{\centering\arraybackslash}X|>{\centering\arraybackslash}X>{\centering\arraybackslash}X} %
    
    \toprule[0.95pt]
    \multirow{2}{*}{Method} & \multirow{2}{*}{Training Source} & \multicolumn{2}{c}{ScanNet20 (20)} \\
     & & f-mIoU & f-mAcc \\
    \midrule[0.6pt]
    
    OpenScene & ScanNet & 57.5 & 72.4 \\
    Mosaic3D & \scannet & 65.0 & 82.5   \\
    PLA & \scannet & 19.1 & 41.5 \\
    RegionPLC & \scannet & 55.6 & 76.3 \\
    OV3D & \scannet & 64.0 & 76.3 \\
    \rowcolor{gray!15}
    \ours & \scannet & 35.4 & 57.9  \\
    \bottomrule[0.95pt]
    \end{tabularx}
    
    \vspace{-2mm}
    \caption{\textbf{Zero-Shot 3D Semantic Segmentation on ScanNet20 Benchmark.} The results for the baselines are taken from~\cite{lee2025mosaic3d}. Ours observes many faulty predictions and have poor performance, we identify the issue in the inconsistency during 2D feature map collection when labeling Gaussians, which leads to corrupted 3DGS-feature pairs.}
    
    \label{tab:scannet20_semseg}
\end{table}

\subsection{3DGS Self-Supervised Pretraining}

\begin{table}[!t]
    \centering
    \footnotesize
    \begin{tabularx}{\columnwidth}{l|>{\centering\arraybackslash}X>{\centering\arraybackslash}X|>{\centering\arraybackslash}X>{\centering\arraybackslash}X}
        \toprule[0.95pt]
        \multirow{2}{*}{Method} & \multicolumn{2}{c}{\scannetpp 2D } & \multicolumn{2}{c}{\scannetpp 3D}   \\
         & L2 & cosine  & L2 & cosine  \\
        \midrule[0.6pt]
        VAE16 layer5  & 0.008 &  0.182  & 0.005 &  0.028 \\
        VAE16 layer6  & 0.008 &  0.172  & 0.005 &  0.023 \\
        VAE64 layer5  & 0.007 &  0.081  & 0.004 &  0.014 \\
        VAE64 layer6  & 0.007 &  0.079  & 0.004 & 0.013 \\
        \bottomrule[0.95pt]
    \end{tabularx}
    
    \vspace{-2mm}
    \caption{
        \textbf{AutoEncoder Ablation Experiments.} We report the feature compression performance with $\mathcal{L}_2$ loss and cosine similarity on unseen 3D language label. We ablate on different autoencoder architectures and training sources.
    }   
    \label{tab:autoencoder_abaltion}
\end{table}

In \cref{tab:autoencoder_abaltion}, we present an ablation study that compares various autoencoder architectures trained on the ScanNet++ training set and evaluated on the validation set. In the “\scannetpp 2D” columns, the autoencoder is trained using SigLIP2 features preprocessed from images, while in the “\scannetpp 3D” columns, it is trained directly on Gaussian SigLIP2 features. At inference, we measure both the L2 distance and cosine similarity on 3D Gaussian SigLIP2 features. The results indicate that training the autoencoder directly on 3D data yields notably better performance, whereas increasing the network depth from layer5 to layer6 leads to only marginal improvements. Moreover, the 64-dimensional latent space consistently outperforms the 16-dimensional counterpart. Consequently, as detailed in \cref{supp:sub:ae}, we adopt the 3D-trained autoencoder with layer5 and a 16-dimensional latent space as our default configuration, striking a favorable balance between efficiency and accuracy.

\begin{table}[!t]
    \centering
    \footnotesize
    \begin{tabularx}{\columnwidth}{l|>{\centering\arraybackslash}X>{\centering\arraybackslash}X}
        \toprule[0.95pt]
        \multirow{2}{*}{Method} & \multicolumn{2}{c}{ScanNet20 (20)}  \\
         & mIoU & mAcc  \\
        \midrule[0.6pt]
        LA16  & \textbf{76.3} &  \textbf{83.9}  \\
        LA64  & 75.8 & 82.2   \\
        LA16 (MGM)  & 76.2 & 83.7  \\
        \bottomrule[0.95pt]
    \end{tabularx}
    
    \vspace{-2mm}
    \caption{
        \textbf{Language Alignment Loss Ablation.} We conduct an ablation study on low-dimensional language feature compression and Masked Gaussian Modeling (MGM)-based pretraining to evaluate their impact on semantic segmentation performance.
    }   
    \label{tab:language_alignment_ablate}
\end{table}

We evaluate the impact of language feature alignment loss dimensionality (64 vs.16) when pretraining exclusively on ScanNet++ and testing on the ScanNet20 benchmark. As shown in \cref{tab:language_alignment_ablate}, we implement two 3-layer MLP architectures: one with uniform dimensions (64 → 64 → 64) for 64 dimension language feature and another with progressively reduced dimensions (64 → 32 → 16). We observe that reducing the latent dimension from 64D to 16D improves mIoU by +\(0.5\%\) in mIoU and +0.7\% in mAcc. This suggests that lower-dimensional language features with hierarchical dimensionality reduction preserve critical indoor scene semantic information. 

\begin{figure*}[t]
\centering
\includegraphics[width=\textwidth]{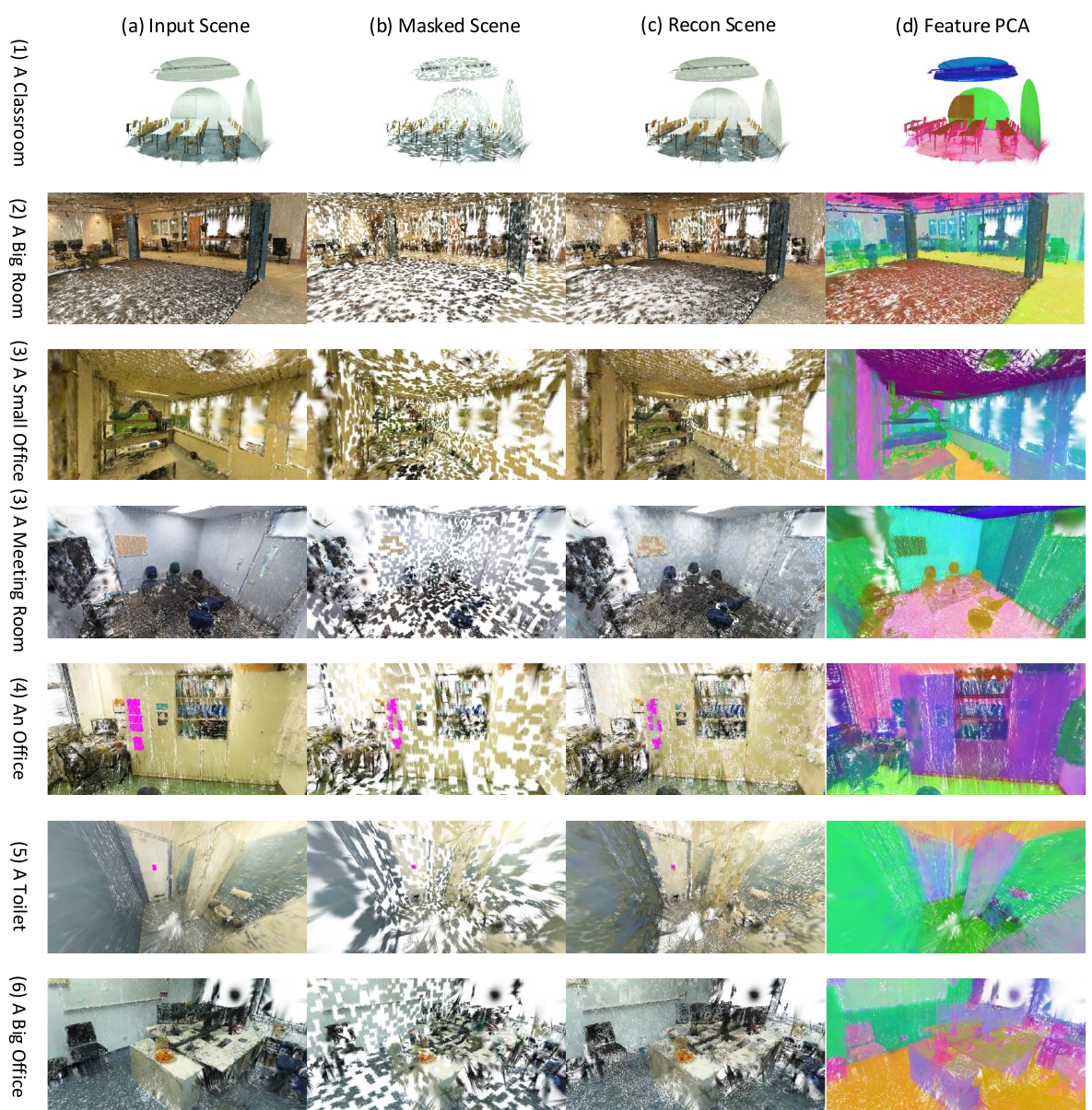}

\caption{\textbf{Self-supervised Reconstructions across Multiple Scenes.} Each row shows (left to right) the unmasked input, masked scene, reconstruction, and a PCA projection of features}
\label{fig:supp_mae_recon}
\end{figure*}

We visualize the pretraining results using Masked Gaussian Modeling (MGM) and language feature alignment loss in \cref{fig:supp_mae_recon}. The model effectively reconstructs masked Gaussian parameters (\eg, position, scale, and rotation) and predicts semantically meaningful language-aligned features for indoor scene understanding.

\cref{tab:MGM_ablation} evaluates how masking ratio and grid size in Masked Gaussian Modeling (MGM) affect downstream semantic segmentation performance. Key observations include: (1) small masking ratios (0.4) degrade mIoU by \(0.9\%\) due to insufficient context learning compared to the ratio 0.6; (2) using a larger grid size $0.15m$ for masking reduces fine-grained detail recovery, leading to a $0.5\%$ drop in mIoU compared to a $0.05m$ grid size.

\begin{table}[!t]
    \centering
    \footnotesize
    \begin{tabularx}{\columnwidth}{l|>{\centering\arraybackslash}X>{\centering\arraybackslash}X|>{\centering\arraybackslash}X|>{\centering\arraybackslash}X>{\centering\arraybackslash}X}
        \toprule[0.95pt]
        \multirow{2}{*}{Mask Ratio} & \multicolumn{2}{c|}{ScanNet20} & Mask &  \multicolumn{2}{c}{ScanNet20}   \\
         & mIoU & mAcc &  Size &  mIoU & mAcc   \\
        \midrule[0.6pt]
        0.4  & 76.1 &  83.6 & 0.02 & 76.9 & \textbf{84.5} \\
        0.5  & 76.4 & 84.1 & 0.05 & \textbf{77.0} & \textbf{84.5} \\
        0.6  & \textbf{77.0} & \textbf{84.4} & 0.10 & 76.6 & 83.8 \\
        0.7  & 76.7 & 83.8 & 0.15 & 76.5 & 84.1 \\
        \bottomrule[0.95pt]
    \end{tabularx}
    
    \vspace{-2mm}
    \caption{
        \textbf{Masked Gaussian Modeling Ablation Experiment.} We analyze the impact of masking ratios and grid sizes in Masked Gaussian Modeling (MGM) on semantic segmentation performance using the ScanNet20 benchmark.
    }   
    
\label{tab:MGM_ablation}
\end{table}

\begin{table}[!t]
    \centering
    \footnotesize
    \begin{tabularx}{\columnwidth}{l|>{\centering\arraybackslash}X>{\centering\arraybackslash}X|>{\centering\arraybackslash}X>{\centering\arraybackslash}X}
        \toprule[0.95pt]
        \multirow{1}{*}{Method} & \multicolumn{2}{c|}{ModelNet10 (10)} & \multicolumn{2}{c}{Omniobject3D (83)}  \\
         & Linear & MLP-3 & Linear & MLP-3  \\
        \midrule[0.6pt]
        MGM  & \underline{77.3} & \underline{83.1}  & \textbf{50.3} & \textbf{57.5}  \\
        +DINO & 74.5 & 80.2 & 40.5 & 42.4     \\
        +iBOT  & 65.2 & 73.8 & 38.4 & 40.8    \\
        +LA & 68.1 & 74.8 & 26.5  & 31.2    \\
        MGM+LA  & \textbf{83.1} & \textbf{84.6} & \underline{47.3}  & \underline{53.2}    \\
        \bottomrule[0.95pt]
    \end{tabularx}
    
    \vspace{-2mm}
    \caption{
        \textbf{Cross Domain Linear Probe Experiments.} We report the mAcc results in the ablation of GaussianSSL on object-level classification tasks using the linear probe.
    }   
    
\label{tab:object_linear_prob}
\end{table}

To evaluate global scene understanding, we design a cross-domain classification task (\cref{tab:object_linear_prob}) using object-level Gaussian splats from~\cite{ma2024shapesplat,ma2024implicitzoolargescaledatasetneural}. We freeze the encoder and embedding layers and train only (1) a linear probe (final layer) to map the features to logits. (2) a 3-layer MLP (512 → 256 → classes) for nonlinear evaluation. We find that progressively adding DINO, iBOT, and language alignment (LA) losses degrades the classification accuracy. This misaligned trend with scene-level benchmarks stems from domain gaps, where the model must map distinct objects (\eg, fridge, oven) to similar scene-level semantics (\eg, "kitchen"). Using masked pretraining and language alignment loss achieves the best performance on ModelNet10 (furniture), whereas MGM alone excels on OmniObject3D (common objects). For ModelNet10, MGM achieves the best 80\% accuracy, whereas for the challenging object dataset (\eg, OmniObject3D), it peaks at ~50\%.

\begin{figure*}[t]
    \centering
    \begin{tabular}{cc}
    \includegraphics[width=\linewidth]{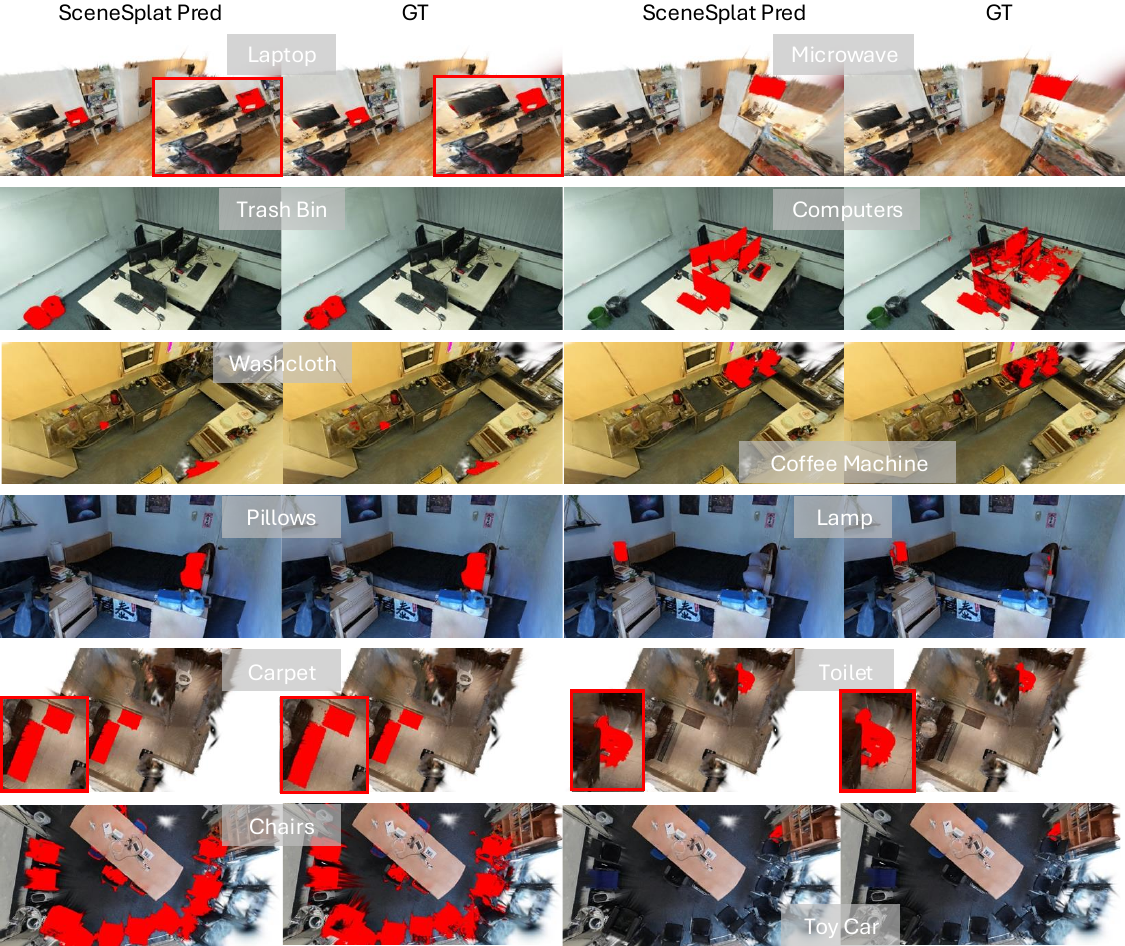}
    \end{tabular}
    \vspace{-1mm}
    \caption{
        \textbf{Comparison of Scene Query Results Using Our Predictions and GT Language Labels on \scannetpp.} 
    }
    \label{fig:scannetpp_query}
\end{figure*}

\section{Datasets Curation and Statistics}
\label{supp:statistics}
\subsection{Analysis}
Gaussian Splatting (GS) has demonstrated strong reconstruction capabilities, but its performance is heavily dependent on raw dataset quality and input conditions. We analyzed both quantitative metrics and visualization results of reconstructed scenes and identified several common failure cases. The primary issues include holes, floating artifacts, and non-smooth surfaces. Through our analysis, we categorize the root causes into four main factors.

\begin{figure}[t]
\centering
\includegraphics[width=0.48\textwidth]{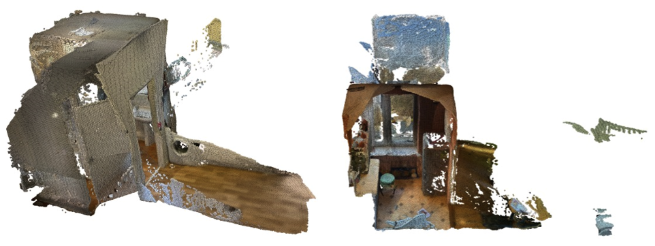}
\caption{\textbf{Example Incomplete Scenes in ARKitScenes.}}
\label{supp:incomplete_ARKitScenes}
\end{figure}

\noindent\textbf{Lack of Frames and Filming Angles.} 
Scenes with fewer than 400 RGB frames often fail to capture a complete representation of the environment. Limited filming angles result in missing perspectives, leading to incomplete indoor structures. \cref{supp:incomplete_ARKitScenes} from the ARKitScenes dataset show partial room reconstructions where corners or ceiling details are lost.

\noindent\textbf{Motion Blur and Camera Instability.}
As seen in some blurry input examples \cref{supp:blurry} in the 3RScan dataset, the rapid, unsteady filming leads to edge deformation, ghosting, and smeared textures, and in unrealistic reconstructions, the sharp edges are lost, significantly reducing the fine details in the rendered GS.

\begin{figure}[t]
\centering
\includegraphics[width=0.48\textwidth]{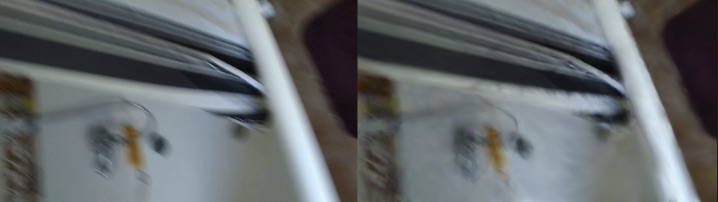}
\includegraphics[width=0.48\textwidth]{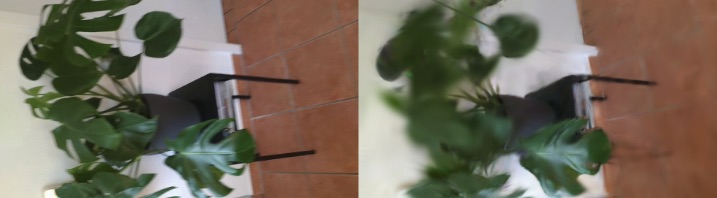}
\caption{\textbf{Blurry Scenes from 3RScan.}}
\label{supp:blurry}
\end{figure}

\noindent\textbf{Indoor-outdoor Lighting Changes.}
Scenes with dynamic indoor-outdoor lighting changes, such as the case from the Hypersim dataset, exhibit severe blurring and loss of surface textures. Large glass ceilings, skylights, and reflective floors further disrupt GS training, as the algorithm struggles with light inconsistency and high contrast between illuminated and shadowed areas. 

\noindent\textbf{Challenges with Transparent and Glass Objects.}
Transparent and glass objects pose a unique challenge, as GS often fails to accurately capture windows or furniture glass, resulting in missing elements or floating artifacts. This issue arises from the inherent difficulty in rendering transparency, where reflection and refraction introduce complexity beyond the current GS capabilities.

\subsection{Visualization}
The \ourdata dataset achieves an impressive average PSNR of 28.17 dB through all the different sources. We provide visualizations that showcase the photorealistic appearance rendering. See \cref{fig:3dgs_renders}.

\begin{figure*}[b]
\ContinuedFloat
    \centering
    \begin{subfigure}[b]{0.48\linewidth}
        \includegraphics[width=\linewidth]{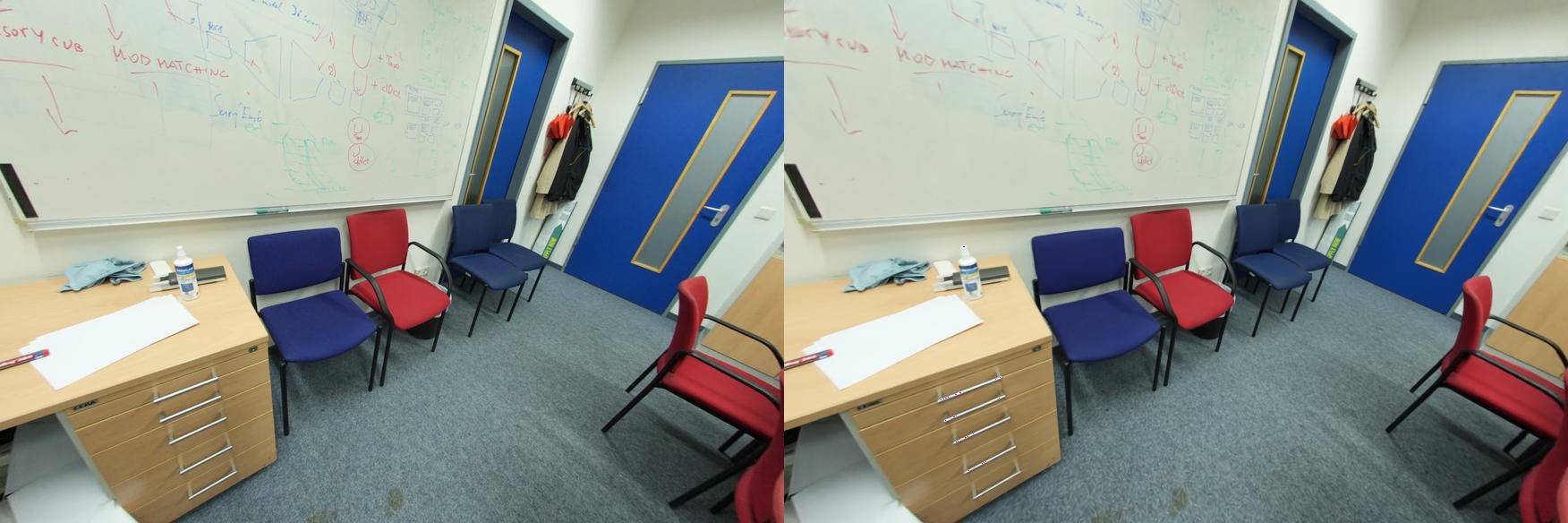}
        \caption{}
    \end{subfigure}
    \begin{subfigure}[b]{0.48\linewidth}
        \includegraphics[width=\linewidth]{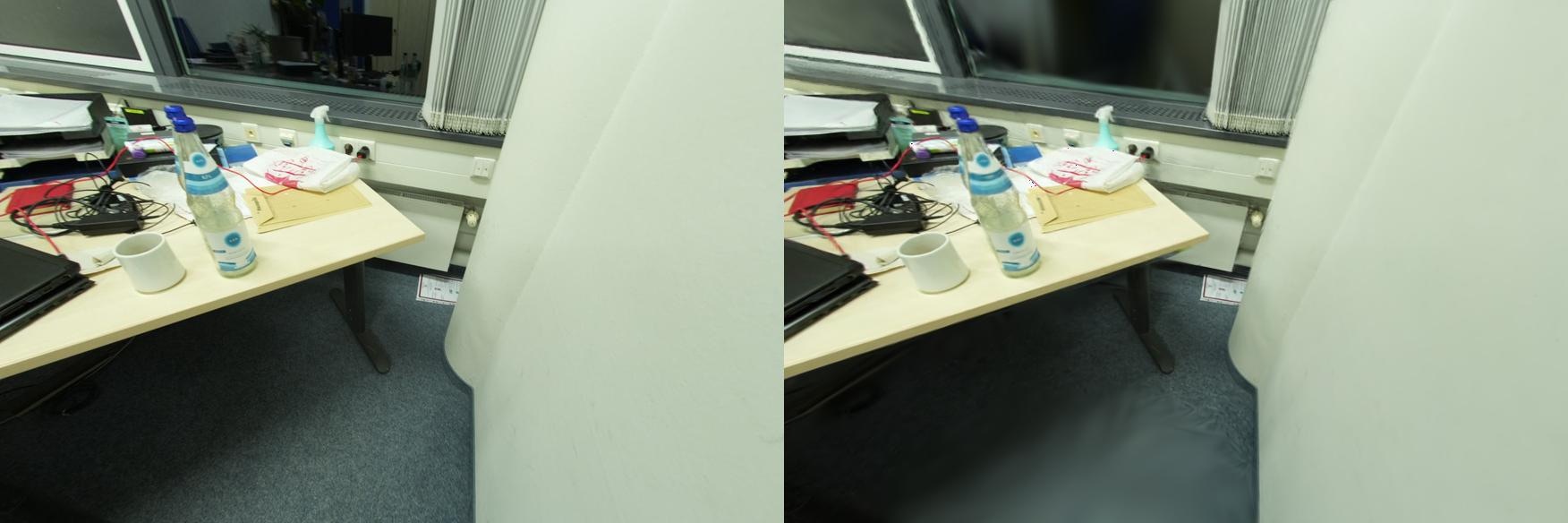}
        \caption{}
    \end{subfigure}
    \begin{subfigure}[b]{0.48\linewidth}
        \includegraphics[width=\linewidth]{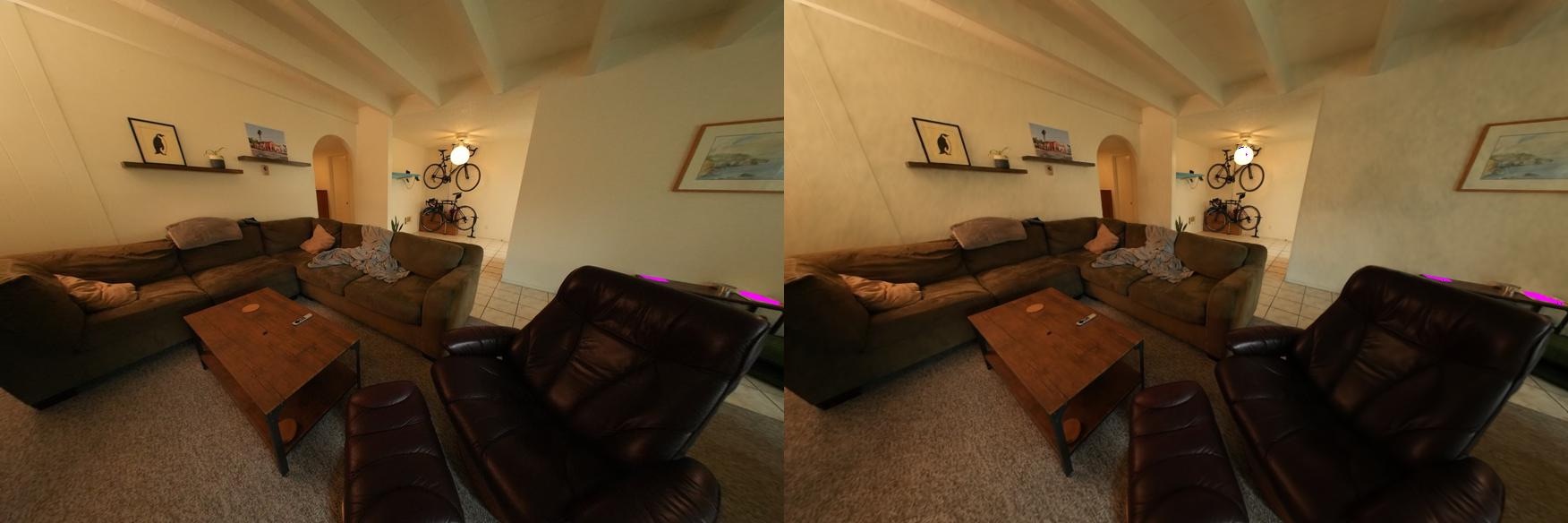}
        \caption{}
    \end{subfigure}
    \begin{subfigure}[b]{0.48\linewidth}
        \includegraphics[width=\linewidth]{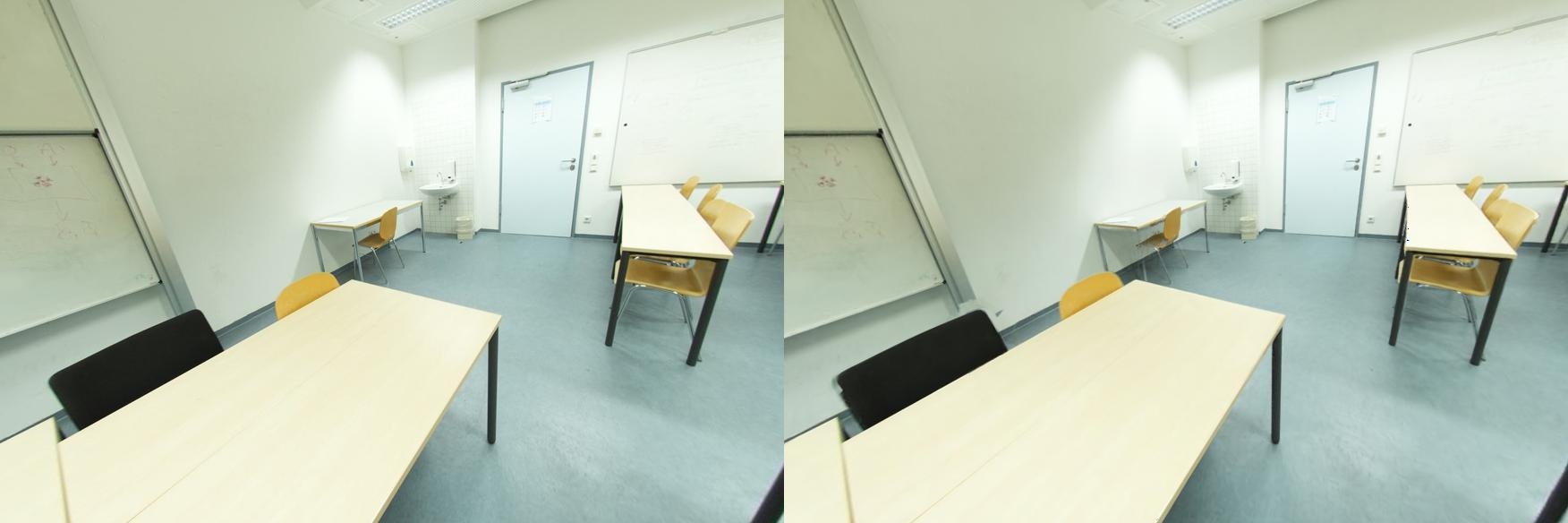}
        \caption{}
    \end{subfigure}
    \begin{subfigure}[b]{0.48\linewidth}
        \includegraphics[width=\linewidth]{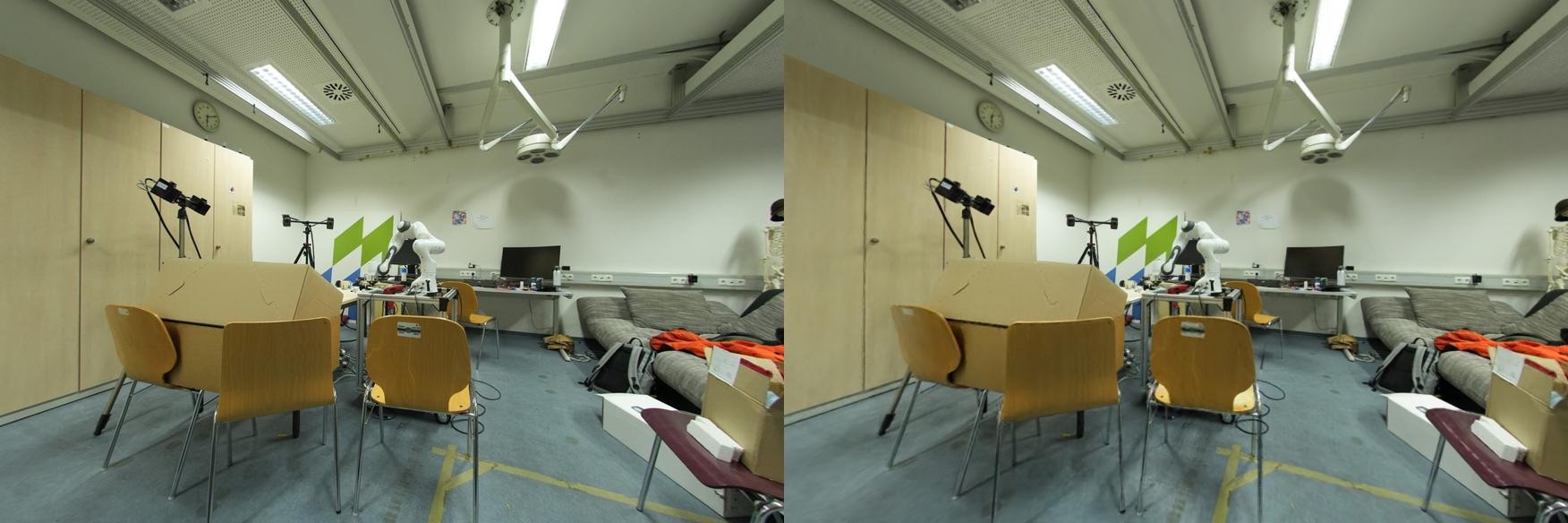}
        \caption{}
    \end{subfigure}
    \begin{subfigure}[b]{0.48\linewidth}
        \includegraphics[width=\linewidth]{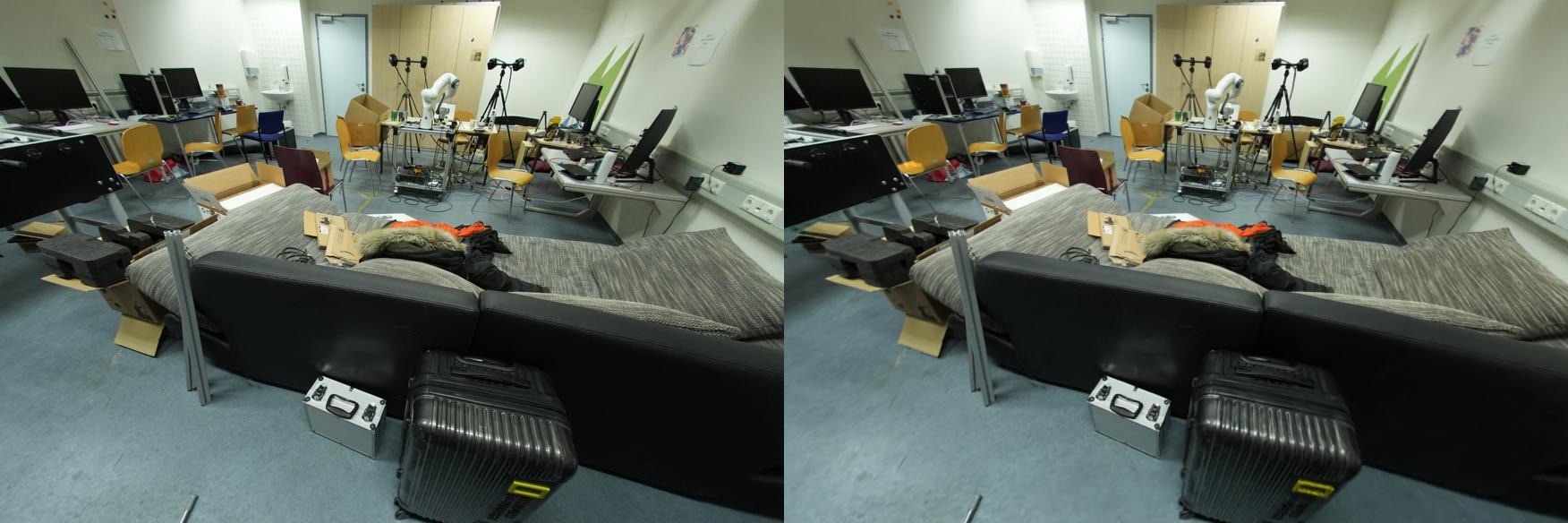}
        \caption{}
    \end{subfigure}
    \caption*{\textbf{\scannetpp GS.} Ground truth (left) and 3DGS rendering results (right).}
    \label{fig:3dgs_renders}
\end{figure*}

\begin{figure*}[t]
\ContinuedFloat
    \centering
    \begin{subfigure}[b]{0.48\linewidth}
        \includegraphics[width=\linewidth]{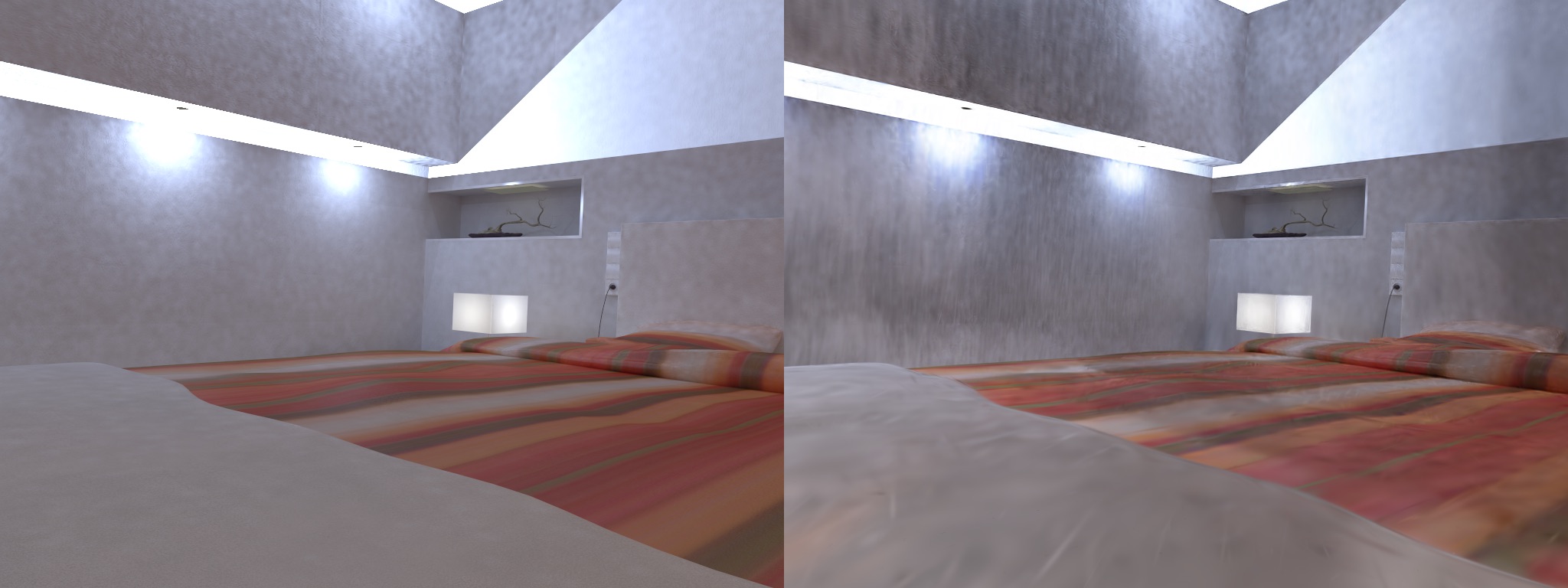}
        \caption{}
    \end{subfigure}
    \begin{subfigure}[b]{0.48\linewidth}
        \includegraphics[width=\linewidth]{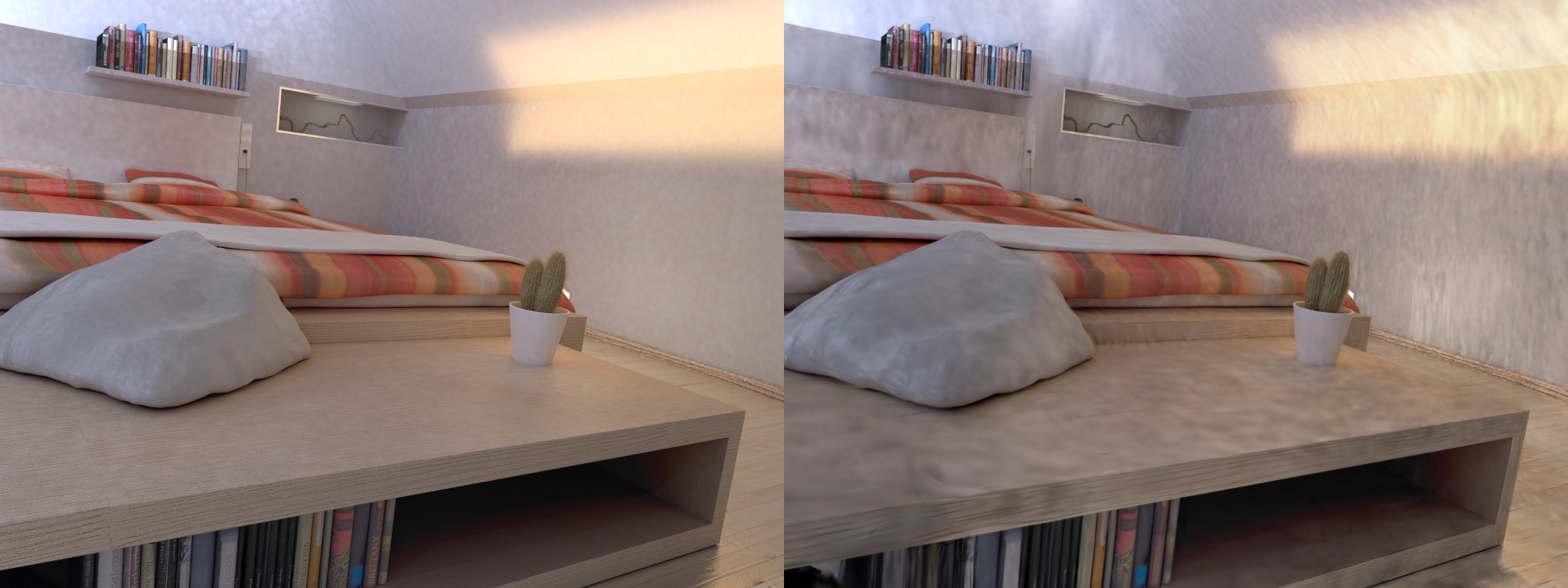}
        \caption{}
    \end{subfigure}
    \begin{subfigure}[b]{0.48\linewidth}
        \includegraphics[width=\linewidth]{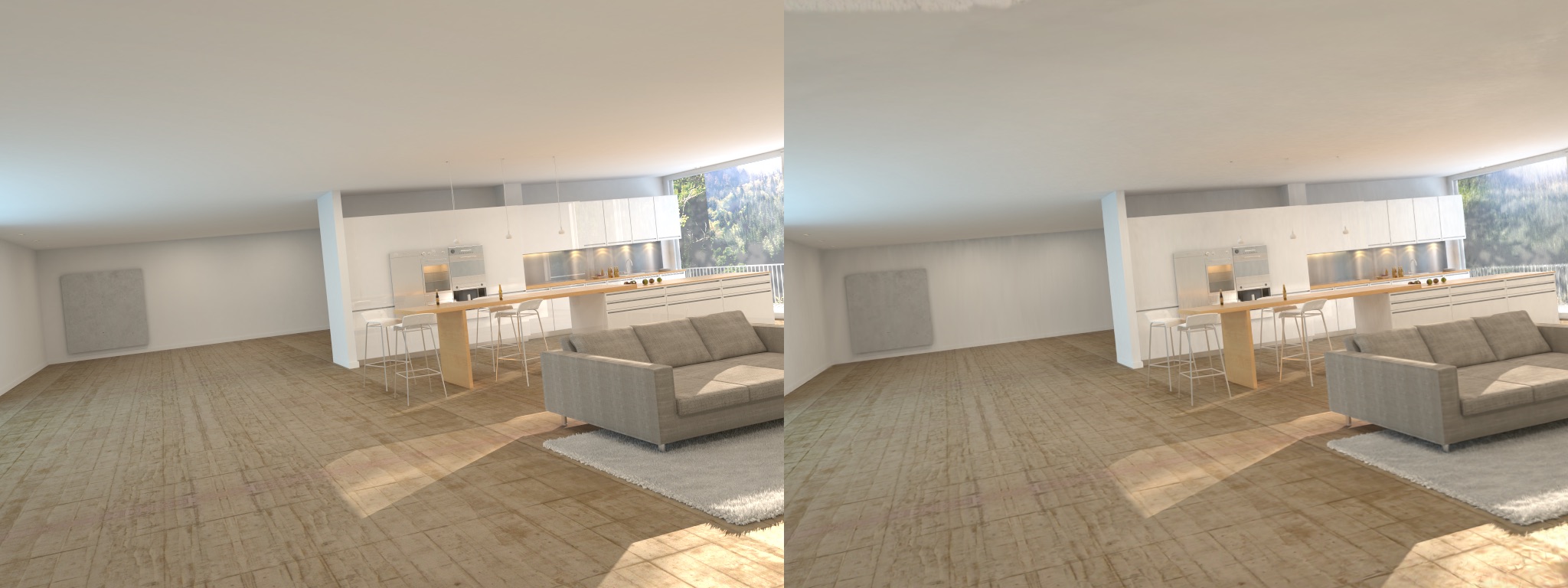}
        \caption{}
    \end{subfigure}
    \begin{subfigure}[b]{0.48\linewidth}
        \includegraphics[width=\linewidth]{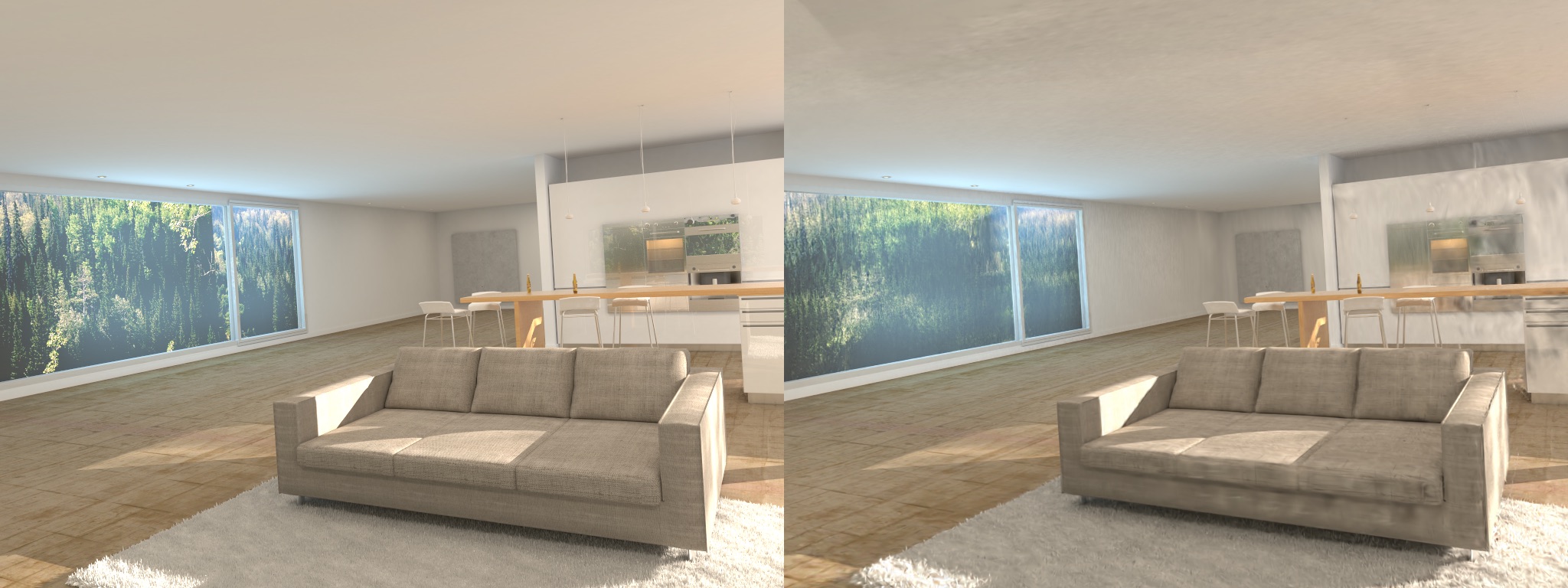}
        \caption{}
    \end{subfigure}
    \begin{subfigure}[b]{0.48\linewidth}
        \includegraphics[width=\linewidth]{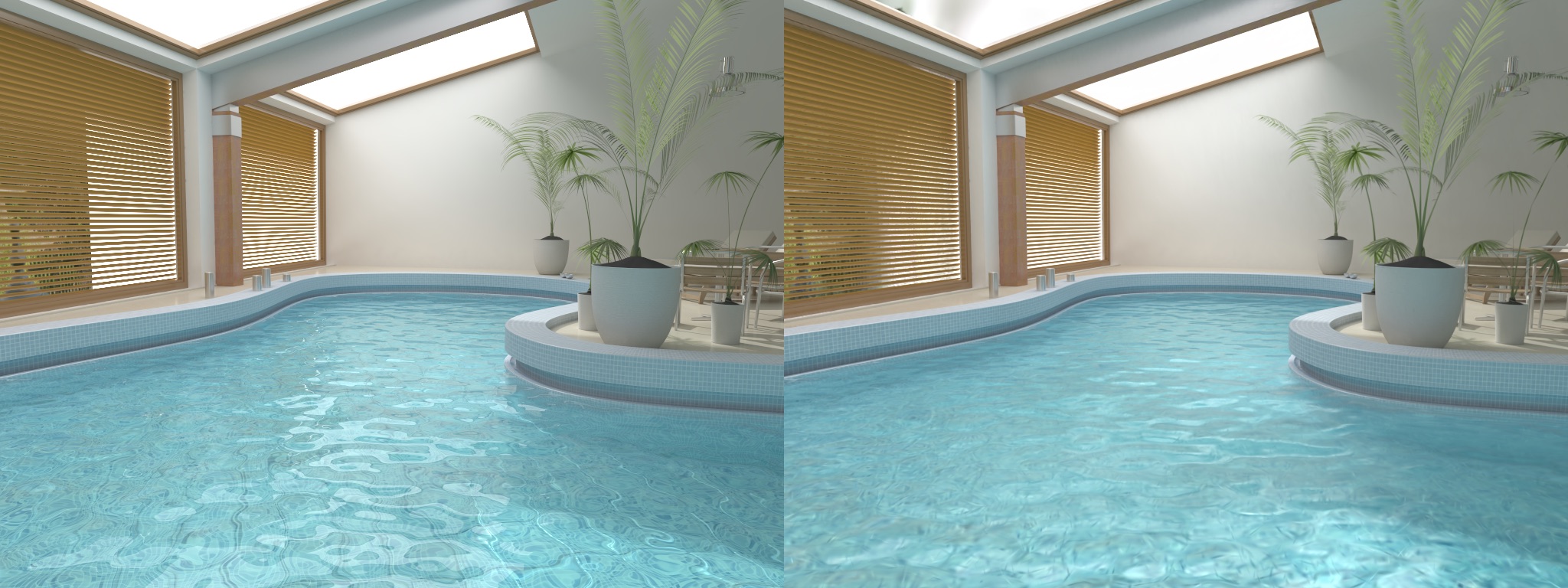}
        \caption{}
    \end{subfigure}
    \begin{subfigure}[b]{0.48\linewidth}
        \includegraphics[width=\linewidth]{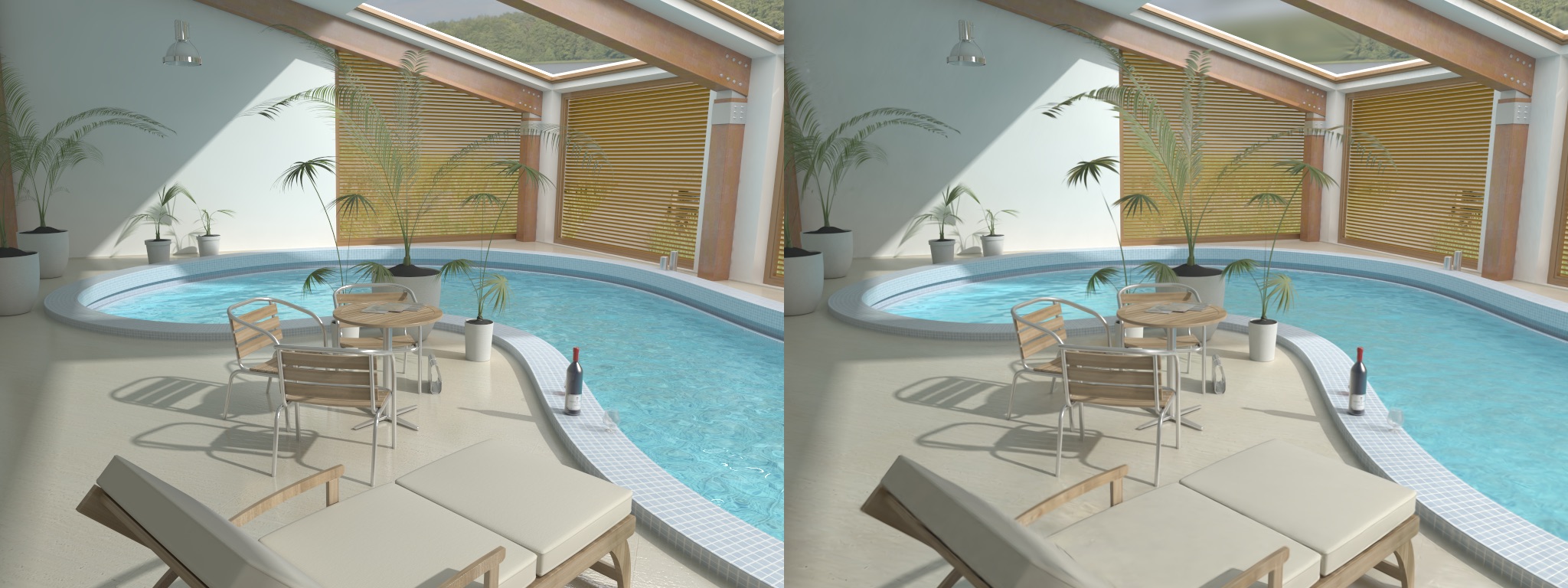}
        \caption{}
    \end{subfigure}
    \caption*{\textbf{Hypersim GS.} Ground truth (left) and 3DGS rendering results (right).}
\end{figure*}

\begin{figure*}[t]
\ContinuedFloat
    \centering
    \begin{subfigure}[b]{0.48\linewidth}
        \includegraphics[width=\linewidth]{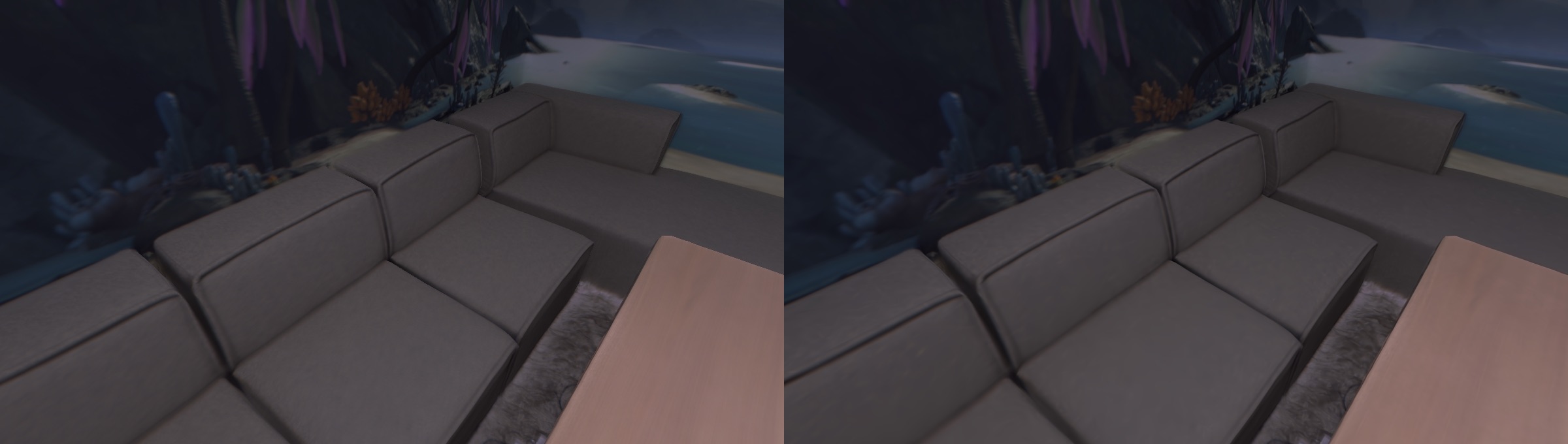}
        \caption{}
    \end{subfigure}
    \begin{subfigure}[b]{0.48\linewidth}
        \includegraphics[width=\linewidth]{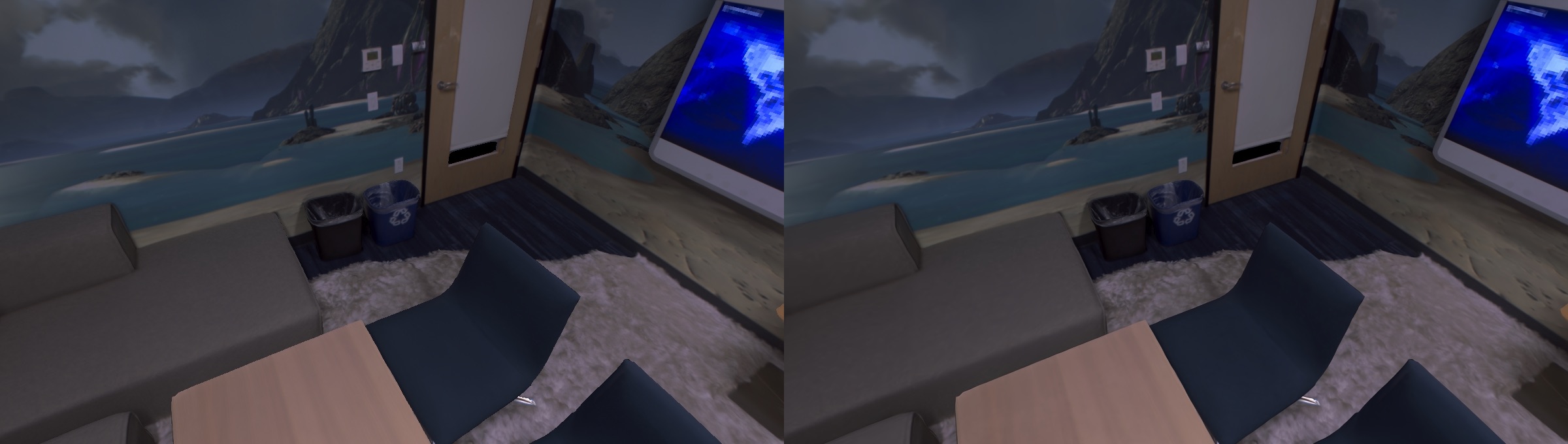}
        \caption{}
    \end{subfigure}

    \begin{subfigure}[b]{0.48\linewidth}
        \includegraphics[width=\linewidth]{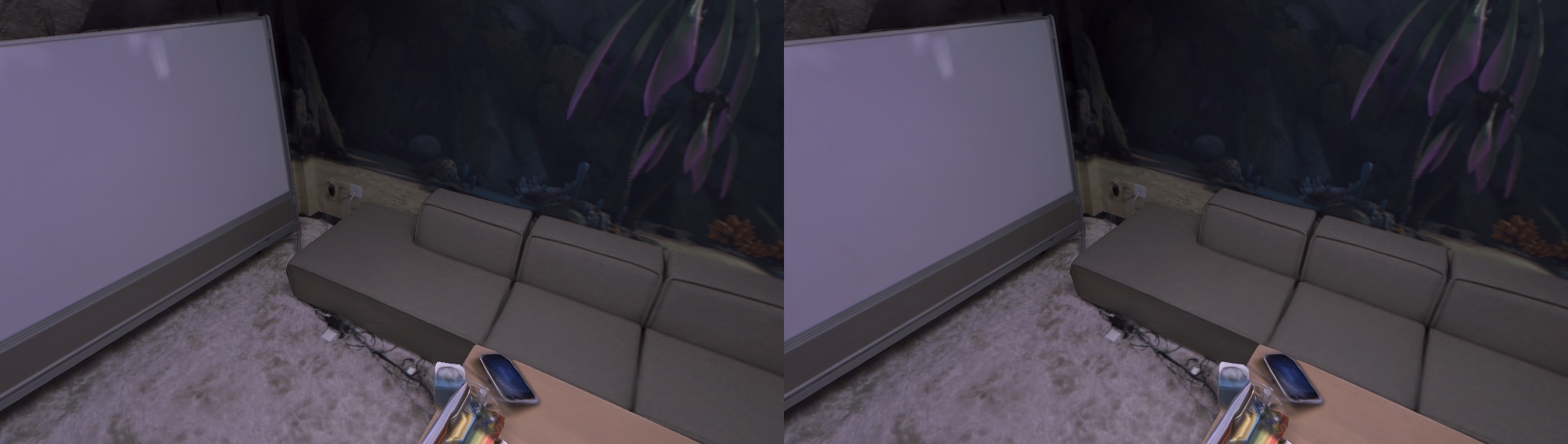}
        \caption{}
    \end{subfigure}
    \begin{subfigure}[b]{0.48\linewidth}
        \includegraphics[width=\linewidth]{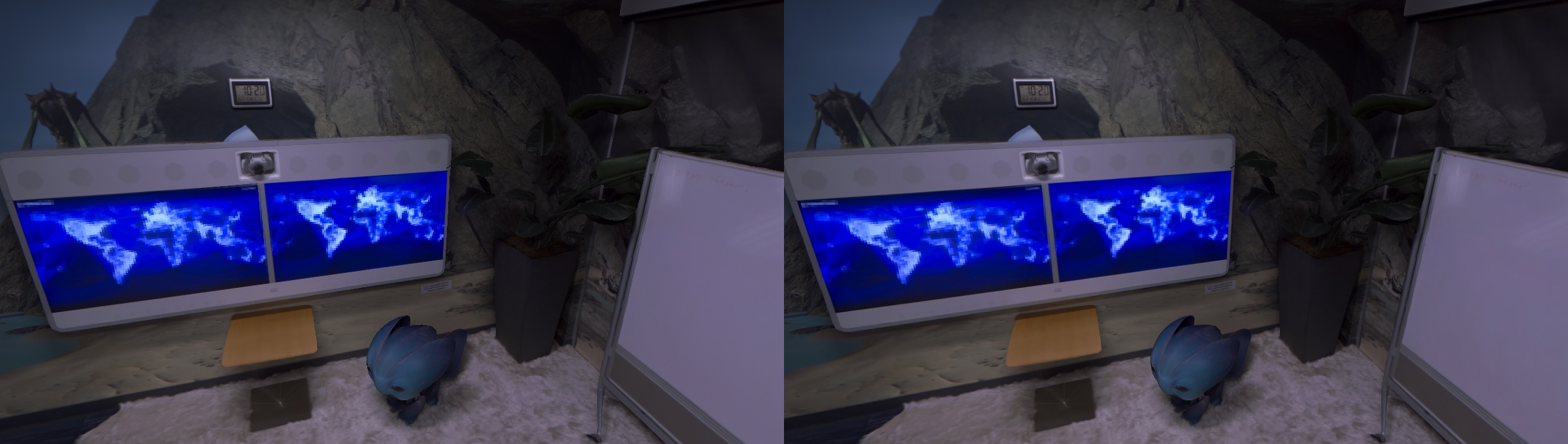}
        \caption{}
    \end{subfigure}
    \caption*{\textbf{Replica Office0 GS.} Ground truth (left) and 3DGS rendering results (right). PSNR: 45.55 dB.}
\end{figure*}

\begin{figure*}[t]
\ContinuedFloat
    \centering
    \begin{subfigure}[b]{0.48\linewidth}
        \includegraphics[width=\linewidth]{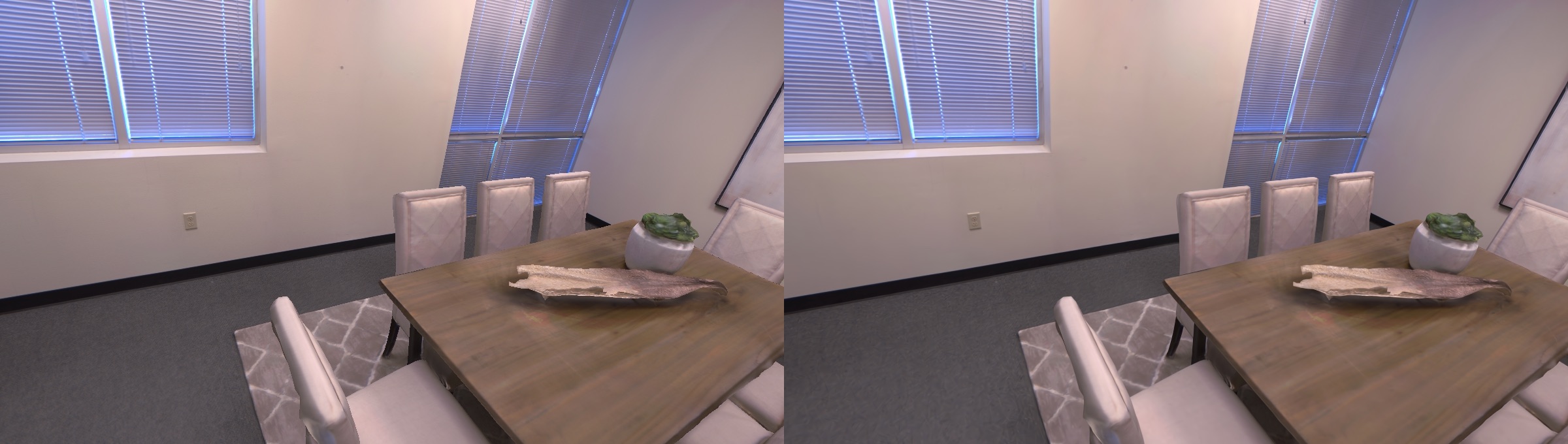}
        \caption{}
    \end{subfigure}
    \begin{subfigure}[b]{0.48\linewidth}
        \includegraphics[width=\linewidth]{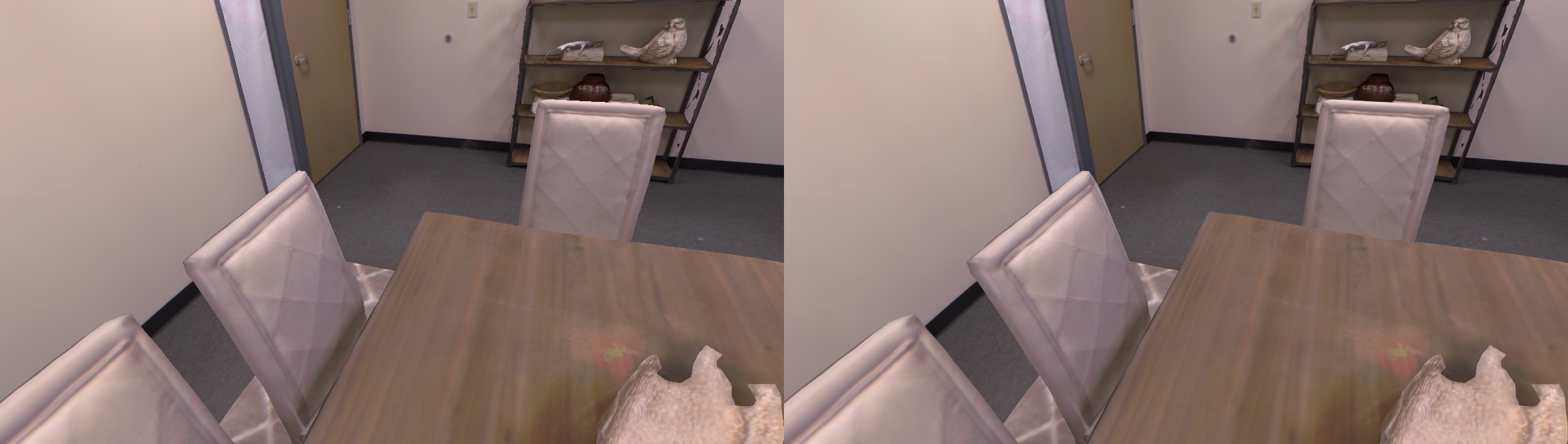}
        \caption{}
    \end{subfigure}
    \begin{subfigure}[b]{0.48\linewidth}
        \includegraphics[width=\linewidth]{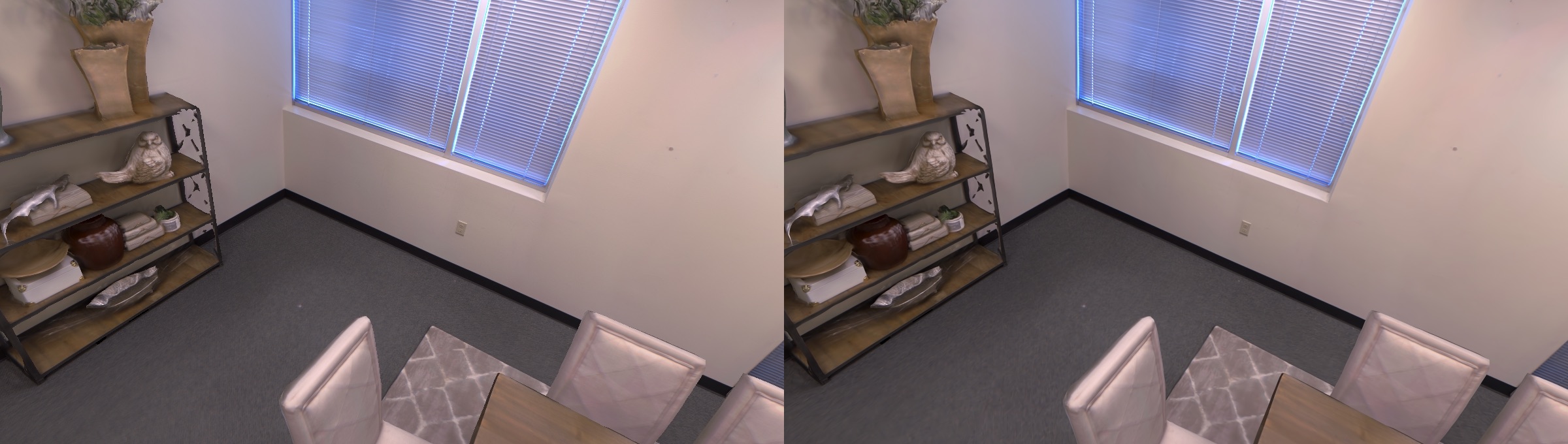}
        \caption{}
    \end{subfigure}
    \begin{subfigure}[b]{0.48\linewidth}
        \includegraphics[width=\linewidth]{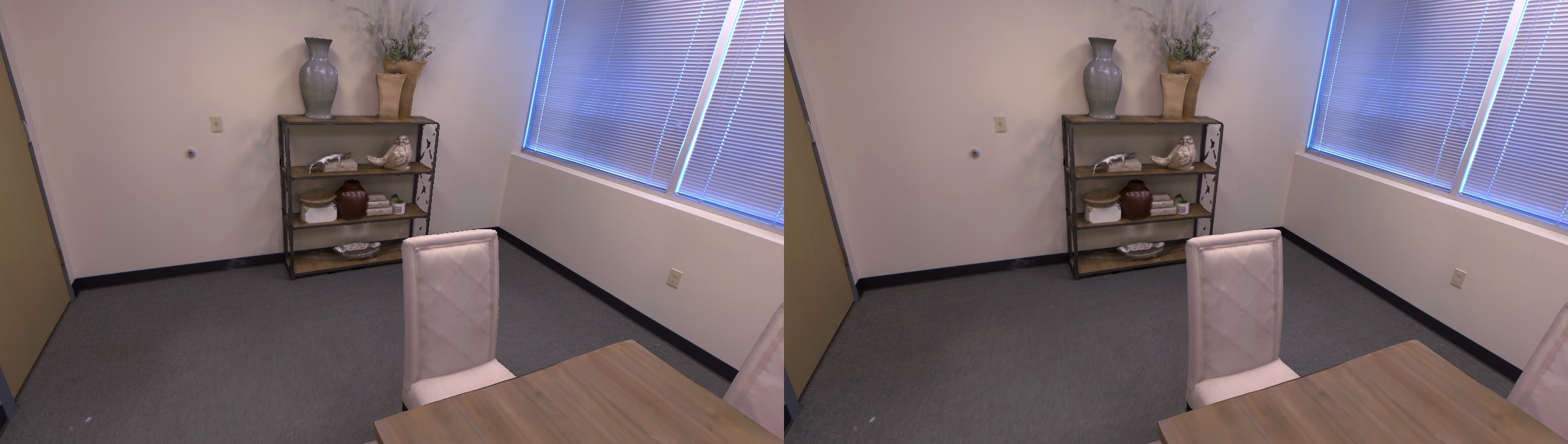}
        \caption{}
    \end{subfigure}
    \caption*{\textbf{Replica Room2 GS.} Ground truth (left) and 3DGS rendering results (right). PSNR: 41.57 dB.}
    \label{fig:replica_3dgs_render}
\end{figure*}

\begin{figure*}[t]
\ContinuedFloat
    \centering
    \begin{subfigure}[b]{0.48\linewidth}
        \includegraphics[width=\linewidth]{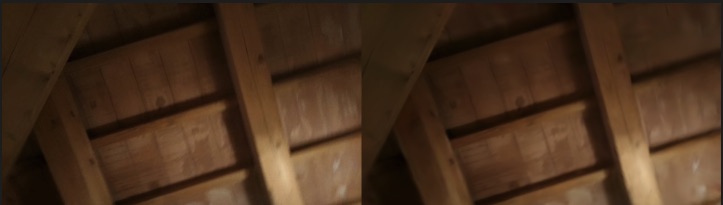}
    \end{subfigure}
    \begin{subfigure}[b]{0.48\linewidth}
        \includegraphics[width=\linewidth]{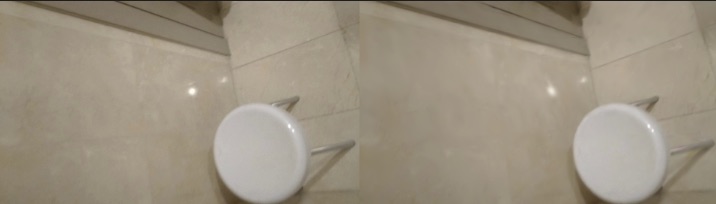}
    \end{subfigure}
    \begin{subfigure}[b]{0.48\linewidth}
        \includegraphics[width=\linewidth]{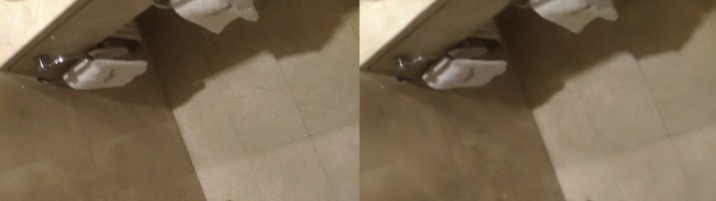}
    \end{subfigure}
    \begin{subfigure}[b]{0.48\linewidth}
        \includegraphics[width=\linewidth]{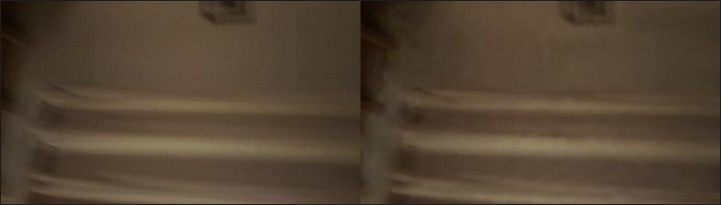}
    \end{subfigure}
    \caption{\textbf{3RScan GS.} Ground truth (left) and 3DGS rendering results (right). PSNR: 34.43 dB.}
    \label{fig:3rscan_3dgs_render}
\end{figure*}

\section{Dataset License}\label{licence}
\ourdata is constructed from several established indoor datasets, each attached with a specific license:
ARKitScenes~\cite{baruch2021arkitscenes} (Apple Open Source License),
Replica~\cite{straub2019replica} (Replica Research License),
ScanNet~\cite{dai2017scannet} (ScanNet Terms of Use),
ScanNet++~\cite{yeshwanth2023scannet++} (ScanNet++ Terms of Use),
Hypersim~\cite{roberts2021hypersim} (CC BY-SA 3.0),
3RScan~\cite{wald2019rio} (3RScan Terms of Use),
and Matterport3D~\cite{chang2017matterport3d} (Matterport Academic License Agreement).
We have carefully structured our distribution approach to respect all original licenses while making our dataset accessible to the research community.
For sources that permit redistribution, we plan to release our data on the Hugging Face under their respective terms of use.
For datasets that require special permission, we co-host the data only after receiving approval from the original teams or let the original dataset team host the 3DGS data.

\section{Limitations} 
The quality of our dataset is largely influenced by the source indoor datasets. Low-resolution and blurry images can cause floating artifacts. Downstream tasks are also limited by the annotations of the original dataset. Temporal inconsistency in the current language label obtaining process needs to be addressed, as it can pollute vision-language pretraining. In addition, we plan to add bounding boxes \cite{irshad2024nerfmae} and language descriptions \cite{mmscan} in the next step.

\section{Impact Statement}
\label{supp:impact}
This work introduces \ourdata, the first large-scale indoor dataset of 3D Gaussian Splatting (3DGS). By providing over 7K annotated scenes, it enables standardized benchmarking for 3DGS-based reasoning in vision tasks. The proposed framework for vision-language pretraining tailored to 3DGS enhances semantic alignment and establishes a clear connection between latent representations and 3DGS scenes. By leveraging knowledge distillation from 2D foundation models, combining contrastive learning and masked gaussian modeling (MGM), our approach significantly outperforms existing methods in 3D semantic segmentation. This study lays the foundation for advancing scalable 3D scene understanding, with broad implications for autonomous systems and augmented reality applications. By publicly releasing the dataset, model, and code, we foster further innovation and facilitate the development of generalizable solutions for 3D scene understanding.

\clearpage
{
    \small
    \bibliographystyle{ieeenat_fullname}
    \bibliography{main}
}

\clearpage
\appendix

\end{document}